%% file: main.tex
\documentclass[10pt,twocolumn,letterpaper]{article}

\usepackage{iccv}
\usepackage{times}
\usepackage{epsfig}
\usepackage{graphicx}
\usepackage{amsmath}
\usepackage{amssymb}

\usepackage{algorithm}
\usepackage{algorithmic}

\usepackage[export]{adjustbox}
\usepackage{makecell}
\usepackage{siunitx}
\usepackage{multirow}

\usepackage{subcaption}
\usepackage{graphicx}
\usepackage{wrapfig}

\usepackage{dsfont}

\input{macros}

\usepackage[pagebackref=true,breaklinks=true,letterpaper=true,colorlinks,bookmarks=false]{hyperref}

\iccvfinalcopy % *** Uncomment this line for the final submission

\def\iccvPaperID{3452} % *** Enter the ICCV Paper ID here

\ificcvfinal\fi

\begin{document}

%%%%%%%%% TITLE
\title{Contact-Aware Retargeting of Skinned Motion}

\author{Ruben Villegas \textsuperscript{\textnormal{1}} 
\qquad Duygu Ceylan \textsuperscript{\textnormal{1}}
\qquad Aaron Hertzmann \textsuperscript{\textnormal{1}}
\qquad Jimei Yang \textsuperscript{\textnormal{1}}
\qquad Jun Saito \textsuperscript{\textnormal{1}}\\
\textsuperscript{\textnormal{1}}Adobe Research \\
}

\twocolumn[{%
\renewcommand\twocolumn[1][]{#1}%
\vspace{-3em}
\maketitle
\vspace{-3em}
\begin{center}
    \centering
    \includegraphics[width=1.0\linewidth]{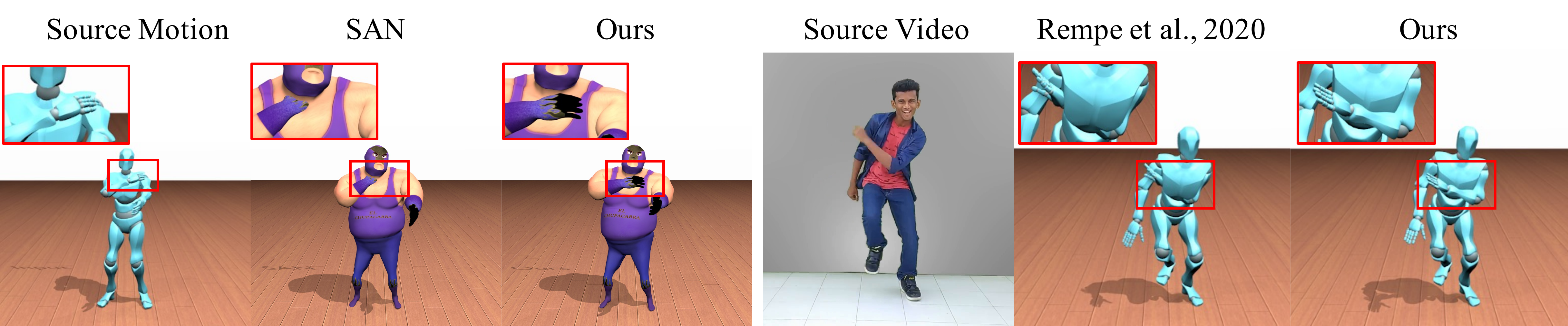}
    \vspace{-.2in}
    \captionof{figure}{Our contact-aware motion retargeting method reduces interpenetration and preserves self-contacts on characters with different geometries in a unified framework while existing methods, like Skeleton-Aware Networks (SAN) \cite{aberman2020skeleton}, do not (Left). In addition, our method generalizes to unseen motions estimated from videos; qualitatively improving noticeable interpenetration in recent human motion estimation methods. We invite readers to watch our \href{https://www.youtube.com/watch?v=qQ4HO2Hibsk}{supplementary video}.}
    \label{teaser}
    %\vspace{-5pt}
\end{center}%
}]

\maketitle
\ificcvfinal\thispagestyle{empty}\fi

% Abstract -----------------------------------------------------------------------------
\input{sections/0_abstract.tex}

% Introduction ------------------------------------------------------------------------------------------------------------------------------------------------------------
\input{sections/1_introduction.tex}

% Related works ------------------------------------------------------------------------------------------------------------------------------------------------------------
\input{sections/2_relatedwork.tex}

% % Method ------------------------------------------------------------------------------------------------------------------------------------------------------------
\input{sections/3_method.tex}

% Experiments ------------------------------------------------------------------------------------------------------------------------------------------------------------
\input{sections/4_experiments.tex}

% Conclusion ------------------------------------------------------------------------------------------------------------------------------------------------------------
\input{sections/5_conclusion.tex}

\input{sections/7_acknowledgements}

{\small
\bibliographystyle{ieee_fullname}
\bibliography{main}
}

% Supplementary material--------------------------------------------------------------------------------------------------------------------------------------------------
\input{sections/6_supplementary}

\end{document}

%% file: macros.tex
\iffalse % don't show comments
    \newcommand\todo[1]{}

    \newcommand{\ruben}[1]{}
    \newcommand{\duygu}[1]{}
    \newcommand{\jimei}[1]{}
    \newcommand{\jun}[1]{}
\else % do not show comments
    \newcommand{\todo}[1]{{\textcolor{red}{[[TODO: {#1}]]}}}

    \newcommand{\ruben}[1]{\textcolor{magenta}{[Ruben: {#1}]}}
    \newcommand{\duygu}[1]{\textcolor{green}{[Duygu: {#1}]}}
    \newcommand{\jimei}[1]{\textcolor{blue}{[Jimei: {#1}]}}
    \newcommand{\jun}[1]{\textcolor{green}{[Jun: {#1}]}}
    
\fi

\iftrue % use space-saving macro

\else % do not use space-saving macro

\fi

\makeatletter

\def\etal{\emph{et al.}}
\makeatother

%% file: sections/0_abstract.tex
\begin{abstract}
This paper introduces a motion retargeting method that preserves self-contacts and prevents interpenetration. Self-contacts, such as when hands touch each other or the torso or the head, are important attributes of human body language and dynamics, yet existing methods do not model or preserve these contacts. Likewise, interpenetration, such as a hand passing into the torso, are a typical artifact of motion estimation methods.
The input to our method is a human motion sequence and a target skeleton and character geometry. The method identifies self-contacts and ground contacts in the input motion, and optimizes the motion to apply to the output skeleton, while preserving these contacts and reducing interpenetration.
We introduce a novel geometry-conditioned recurrent network with an encoder-space optimization strategy that achieves efficient retargeting  while satisfying contact constraints.
In experiments, our results quantitatively outperform previous methods and we conduct a user study where our retargeted motions are rated as higher-quality than those produced by recent works.
We also show our method generalizes to motion estimated from human videos where we improve over previous works that produce noticeable interpenetration.

\end{abstract}
% \vspace{-.2in}

%% file: sections/1_introduction.tex
% \cutsectionup
\section{Introduction}
% \cutsectiondown
% \input{figures/teaser.tex}
Self-contact, where one part of a person's body comes into contact with another part, is crucial to how we perceive human motion. 
% Plausible human motion is critical for various applications including computer games and storytelling in augmented and virtual reality. The way that a character makes contact with itself plays a significant role in how the motion is perceived in terms of plausibility. 
Self-contact often indicates different behaviors or emotional states. For example, people might rest their head in their hands if they are concerned or thoughtful, whereas certain dance moves require one's hands to be placed on one's hips. Conversely, implausible self-contacts, such as hands passing through each other, ruin the physical plausibility of a motion. Hence, handling self-contact is crucial for reconstructing, interpreting, and synthesizing human motion.
% Previous vision and graphics algorithms have not handled self-contact, often leading to invalid interpenetration and failure to preserve important self-contacts.
Note that synthesizing contact for a skinned character requires knowledge of the character's 3D geometry; it cannot be accurately determined from skeleton alone.
%\jun{can we emphasize how the method is mesh-topology-agnostic?}. \aaron{we're not talking about our method yet here}

This paper introduces a motion retargeting algorithm that preserves both self-contact and ground contact, while also reducing interpenetration. Our retargeting algorithm takes an input human motion and a target character,
% \jun{isn't the input "source character, its motion and a target character" (trying to compress skeleton + mesh into a single word character)}
and produces a plausible animation that manifests how that target character might perform the input motion.  Our method first identifies self-contact and foot contacts in the input motion.
We use an energy function that preserves these contacts in the output motion, while reducing interpenetration in the output. %We minimize this energy function via a Recurrent Neural Network (RNN).  
We reason about self-contacts from the character's geometry and foot contacts from the character skeleton, in order to guarantee that contacts will be accurately transferred in the skinned and rendered motion.
Our approach generalizes to any mesh geometry and topology regardless of the number of vertices in the mesh.
% We assume that the input and the output meshes have the same joint angle parameterization, but the skeleton shape and the mesh geometry/topology may be very different between the two characters.

Due to the difficultly of directly optimizing a full motion sequence, we build on previous work and train a Recurrent Neural Network (RNN) to perform retargeting. We find that an RNN does not perfectly satisfies contact constraints given its efficient inference. Therefore, we propose encoder-space optimization, in which we refine the RNN's predictions by iteratively optimizing the hidden units of RNN's encoder.
This process allows our method to efficiently satisfy constraints, by taking advantage of the RNN's smooth, disentangled encoder-space.

Since hands often participate in meaningful self-contacts (e.g., contacts from hands to head, hands to hand, hands to hip), we focus our analysis on contacts between the hands and the rest of the character geometry. Note that our use of geometry means that the output style will depend on character mesh: a bulkier character is more constrained in their movements, which is reflected in our results. We evaluate our method on various complex motion sequences as well as a wide range of character geometries from skinny to bulky. We show that our method provides a good balance between preserving input motion properties while preserving self-contacts and reducing implausible interpenetration. In our qualitative and quantitative experiments, we outperform the state-of-the-art learning-based motion retargeting methods. In addition, we qualitatively show our method generalizes to real scenarios by retargeting motion extracted from human videos. Our method improves upon methods that only consider the skeleton when estimating the motion.

% \ruben{mention virtual characters}

%% file: sections/2_relatedwork.tex
\section{Related Work}

Most early methods for motion retargeting perform a constrained optimization problem over the skeleton of a character. Gleicher \cite{gleicher} optimized the output motion sequence guided by kinematic constraints. Several authors include physical plausibility constraints in the optimization \cite{Popovic:1999:PBM,tak2005physically,al2018robust}, including ground contact constraints. 
Another approach is to first solve an inverse kinematics (IK) problem for each motion frame, and then apply spacetime smoothing to the result \cite{lee1999hierarchical,choi,Unzueta_2008,Aristidou_2011,Harish_2016}.

%out the results by fitting a multi-level B-Spline. Subsequently, Tak and Ko \cite{tak2005physically} proposed a model for physically plausible motion by adding dynamics constraints for sequential filtering in the optimization. Choi and Ko \cite{choi} tackled the motion retargeting problem in an online fashion by performing per-frame IK that solves for the change in joint angles guided by the change in end-effector positions and motion similarity. More recent methods \cite{Unzueta_2008,Aristidou_2011,Harish_2016} focused on improving IK in terms of accuracy and speed.

The above methods either use expensive optimization algorithms or else suboptimal spatiotemporal smoothing. More recently, deep learning methods can resolve both of these issues.
%In recent years, deep learning-based algorithms for motion retargeting have emerged given the existence of large scale datasets such as Mixamo. 
Villegas et al.~\cite{Villegas_2018_CVPR} train a recurrent neural network with a forward-kinematics layer for unsupervised motion retargeting. Lim et al.~\cite{Lim_pmnet_BMVC19} train a feed-forward motion retargeting network that decouples the local pose and the overall character movement. Aberman et al.~\cite{Aberman_2019} proposed a supervised method for 2D motion retargeting that learns disentangled representations of motion, skeleton, and camera view-angle. Aberman et al.~\cite{aberman2020skeleton} proposed Skeleton-Aware Networks (SAN) which learn to retarget between skeletons with different numbers of joints.  However, this method requires retraining for each new set of skeleton topologies, and their method does not take character geometry into account.
% In addition, this method is an offline motion retargeting method (i.e., it must read the full sequence at once) while our method retargets the source motion one frame at time in an online fashion.
% \ruben{They do generalize to short/tall, but there is no geometry being modeled (e.g., bulky / skinny)}%\ruben{Mention that skeleton aware networks don't generalize outside of the learned topologies unless a new network is trained.}
%\aaron{do they generalize to different skeleton geometries (i.e., short/tall?)} \ruben{No. They don't take geometry into account. Someone asked this during SIGGRAPH because they noticed penetrations, and they admited to it. I just mentioned it above.}
Moreover, these network-based retargeting methods cannot precisely satisfy constraints, a limitation we address with encoder-space optimization.

%Different from all these methods, our work focuses on retargeting motion that respects the target character geometry within a unified framework and preserves self-contacts that are key to motion perception. %that can handle any number of vertices within the same neural network architecture and learned weights without having to resource to training any additional networks for different characters.

Only a few older works consider self-contacts in retargeting.  Lyard and Magnenat-Thalmann \cite{LyardAdaptation} use a heuristic sequential optimization strategy.
Ho and Shum \cite{ho_2013} adapt robot motions with a spacetime optimization strategy that prevents self-collisions.
More recently, Basset \etal~\cite{BASSET202011,10.1145/3359566.3360075} proposed a constrained optimization algorithm with attraction and repulsion terms to preserve contacts and avoid interpenetration.
These works, however, require exact vertex correspondences between the input and target meshes, expensive constrained optimization over the entire sequence, and smoothing as a post-process.
Liu \etal~\cite{10.1145/3274247.3274507} requires input/target mesh calibration, and solves for bone, surface smoothness, and volumetric constraints by solving a linear system.
Our method is general to any source/target mesh, retargets motion frame-by-frame, and all mesh and skeleton shape constraints are implicitly handled. 

% We attribute our faster convergence to the smooth,low-dimensional, decoupled embedding learned by our net-work which makes optimization simpler. 
% Our method does not need to model such constraints  to  successfully  retarget  motion  into  the  target character.

% Our method shows how to add these important capabilities to effective motion inference and synthesis algorithms.

%All of above methods use hand-designed constraints specific for the particular tasks, 

% solve the optimization problem by incorporating objectives hand-designed for particular motions along with character based heuristics and constraints. In addition, 
%and these methods retarget motion at the skeleton level while ignoring the character geometry which plays an important role in terms of handling self-contacts and interpenetrations. Moreover, the above methods either perform expensive spacetime optimization, or suboptimal motion smoothing techniques. 
%Our method addresses this limitation by explicitly modeling the geometry of the characters during the retargeting process. %Thus, the resulting motion preserves semantically meaningful motion at the geometry level by avoiding geometry self-intersections and preserving contacts.

3D character geometry has been considered in other animation-related tasks to improve the motion plausibility. Tzionas et al.~\cite{Tzionas:IJCV:2016} incorporated collision detection for hand motion capture and Hasson et al.~\cite{hasson19_obman} used a mesh-level contact loss for jointly 3D reconstructing the hand and the manipulating object. Body geometry is modeled while estimating and synthesizing plausible humans that interact with 3D scene geometry~\cite{PROX:2019,PSI:2019}, and with other characters \cite{SpatialRelationship}. Interpenetration and character collisions with surrounding objects are handled in character posing~\cite{kim_2020}.
However, it is nontrivial to extend these posing methods to retargeting while maintaining the naturalness of the motion.
Our method instead infers character specific geometric constrains (contacts) directly from the mesh and maintains them during retargeting.
In works concurrent to ours, Smith et al.~\cite{Smith_hand_tracking:2020} and Mueller et al.~\cite{Mueller:CVPR:2021} demonstrate the importance of self-contact handling for detailed hand tracking and accurate human pose estimation, respectively.

Recent works have investigated the use of optimization as supervision when ground-truth is not available. Kolotouros et al.~\cite{kolotouros2019spin} proposed a learning scheme for 3D human pose and shape estimation that performs iterative 3D human model fitting which provides weak supervision during learning. Our proposed geometry-aware RNN is trained in a weakly supervised manner using an energy function that preserves self-contacts and ground contacts. In addition, our encoder-space optimization strategy uses the weak supervision provided by our energy function to precisely satisfy constraints at test time.

%% file: sections/3_method.tex
\input{figures/overview}
\section{Contact-Aware Motion Retargeting} \label{sec:training_energy}
% Brief summary of this section
%Our goal of our method is to  in this work is to develop a motion retargeting algorithm that preserves the characteristics of the source motion while avoiding self-penetrations of the target character (Figure \ref{fig:overview}). 

This section describes our motion retargeting method; the main steps are summarized in Figure \ref{fig:overview}.
The inputs to our method are a source motion represented as the skeleton motion $m^A$ and the skinned motion $\hat{v}^A$, and a target character represented by the skeleton $\bar{s}^B$ and the skinned geometry $\bar{v}^B$ in a reference pose (e.g., T-pose). Then, our goal is to output the skeleton motion $m^B$ as well as the corresponding skinned motion $\hat{v}^B$ that preserves the ``content'' of the source motion, including self-contacts and ground contacts, while also not introducing interpenetration. The self-contacts and ground contacts in the source motion are automatically estimated by a contact detection step, described in Section \ref{sec:contact}.
We next describe our energy function, geometry-conditioned RNN, and our encoder-space optimization strategy to satisfy hard constraints.
%The core of our method is an energy function that is composed of two main terms. The first term aims to preserve the contacts in the source motion, and the second prevents self-penetrations of the target character to avoid generating unnatural looking motion. 

%We train a geometry conditioned recurrent neural network (RNN) to minimize this energy function. Our network incorporates a forward kinematics layer and a skinning layer that generates the target skeleton and the skinned motion respectively. In the remaining of this section, we will discuss the details of our energy function (Section \ref{sec:energy}), how we minimize it using an RNN (Section \ref{sec:geocond_rnn}), and provide a procedure to detect and transfer the meaningful contacts in the source motion (Section \ref{sec:contact}). 

\subsection{Energy Function} \label{sec:energy} 
Given the source motion, skeleton, and inferred contacts, we formulate an energy function for the output motion, which includes novel terms to preserve the input contacts, and to reduce the introduction of interpenetration.
%Our motion retargeting method is driven by a novel energy formulation that models the motion both at the geometry and the skeleton level. On the geometry level, our goal is to preserve the meaningful contacts in the source motion while avoiding self-penetrations of the target skin. 
Like previous work, we also include terms to preserve 
%In order to avoid deviating too much from the source motion, we also try to preserve 
the source motion, represented in terms of skeleton motion, i.e., joint rotations, global trajectory, end-effector velocities, and foot contacts. We define the full energy function as:
\begin{align} \label{eq:full}
E_{\mathit{full}}(t) &= E_{\mathit{geo}}(t) + E_{\mathit{skel}}(t),
\end{align}
where $E_{geo}$ is the geometry based motion modeling term focusing on vertex contacts and interpenetration and $E_{skel}$ is the skeleton based motion modeling term. This energy is defined and optimized for each frame $t$ in an online retargeting fashion.
For simplicity, we omit the input $t$ in the remaining of the paper. %Next, we discuss the details of each of these terms.
\vspace{-.1in}
\subsubsection{Geometry-level modeling terms} \label{sec:geometry_modeling}
Modeling motion at the geometry level requires two parallel objectives working together: modeling mesh contacts and reducing interpenetration. If we model the mesh contacts alone, there is nothing to prevent the retargeted motion from perfectly reaching that contact while introducing large penetrations (e.g., left hand going through the torso to touch the right arm). On the other extreme, if we completely avoid interpenetration, meaningful contacts are likely to be lost as these may cause slight interpenetration. Therefore, our geometry modeling energy is defined by:
\begin{equation}
E_{\mathit{geo}} = \lambda E_{\mathit{j2j}} + \beta E_{\mathit{int}} + \gamma E_{\mathit{foot}},
\end{equation}
% where $E_{j2j}$ is our self-contact modeling term, $E_{int}$ is our interpenetration reduction term, and $E_{\mathit{foot}}$ is the foot contact modeling term.
where $E_{j2j}$ is our self-contact term, $E_{int}$ is our interpenetration reduction term, and $E_{\mathit{foot}}$ is the foot contact term.

% \vspace{-.25in}
\paragraph{Self-contact preservation.} 
This step takes as input a set of vertex contact constraints on the output motion for a particular time frame, identified during the contact detection step (Section \ref{sec:contact}). Specifically, the input is a set of  $\mathcal{V} = \{(i,j)\}$ where tuple $(i,j)$ indicates that vertices $v_i$ and $v_j$ of the output mesh should be in contact. 
This is measured as the average distance between such pairs of vertices:
\begin{align}
    E_{\mathit{j2j}} &= \frac{1}{|\mathcal{V}|}\sum_{ (i,j) \in \mathcal{V} }  \| v_i - v_j \|^2
\end{align}
% \aaron{is this still correct? The previous version had weights that depended on the number of other vertices included in each group of points in contact.}
% \ruben{Yes. This is correct. This is actually a more concise version of it that's easier to understand. Before, I had basically all the implementation details which may not be that important to understand the idea.}
%\begin{equation}
%E_{\mathit{j2j}}(t) = \frac{1}{n_t^c} \sum_{i=0}^{n_t^c} \frac{1}{|VC_t^{i}|} \sum_{\hat{v}_t^{j,B},\hat{v}_t^{k,B} \in \ VC_t^{i}} \| \hat{v}_t^{j,B} - \hat{v}_t^{k,B} \|^2_2,
%\end{equation}
%%%duygu here
%where $n_t^c$ is the number of contacts detected at frame $t$ and $VC_t^{i}$ is the set of vertex pairs involved in a particular contact $i$. $VC_t^{i}$ are computed by the correspondence estimation method described above.
%\aaron{there are way way too many subscripts and superscripts here, it's very hard to read. Most of them are unnecessary. Why is $B$ here at all? And, if we define the set of contact constraints as $\{(i,j,t)\} \in \mathcal{V}$, then we can get rid of nearly all the subscripts/superscripts in here and sum over the tuples in $\mathcal{V}$. We could also have one set for each $t$: $\mathcal{V}_t$ if we want to sum over $t$ separately.}

\paragraph{Interpenetration reduction.}
%When retargeting motion between source and target characters that vary in size and shape significantly, e.g., retargeting from a tiny to a bulky character, skeleton-based methods often result in invalid interpenetrations, e.g., the arm passing through the torso. 
We reduce mesh penetrations with a penalty similar to that of Tzionas et al.~\cite{Tzionas:IJCV:2016}. Specifically, at each time step, we detect the set of colliding triangles $\mathcal{F} = \{(r,i)\}$ using a bounding volume hierarchy~\cite{egst.20041028}. We further define a local 3D distance field for each triangle in the form of a cone. For a pair of colliding triangles $f_r$ and $f_i$, the position of one triangle inside the local 3D distance field of the other determines the penetration term. Thus, we define $E_{\mathit{int}}$ as follows:
\begin{align}
\label{eq:int}
    E_{\mathit{int}} = \sum_{(r, i) \in \mathcal{F}}  &\left(\sum_{v_j \in f_r} w_{r,i} \| - \psi_{i}(v_j)n_j \|^2 + \right. \nonumber \\
    & ~~~~\left.\sum_{v_k \in f_i} w_{i,r} \| -\psi_{r}(v_k)n_k\|^2 \right).
\end{align}
where $v_j$'s are vertices in one of the triangles, $n_j$ are the corresponding per-vertex normals, 
and $\psi_i$ and $\psi_r$ are the distance fields for triangles $i$ and $r$. The distance fields are positive for triangles that interpenetrate, and zero otherwise. See Tzionas et al.~\cite{Tzionas:IJCV:2016} for further details.
% \aaron{why are there negative signs inside the square? aren't the normals unit vectors, so this just boils down to $\psi^2$?} \ruben{I simply copied the equations from their paper. I will check if there is any particular reason they do this.} \aaron{I took a look at that paper and I don't understand it there either. (They say $n'n=1$ in the subsequent equation for point-to-plane distances, but not for this equation, but they have the same name... I think this paper is just confusingly written. For our equation here, I think you can just write what we actually did. If it reduces to the square of the distance field, we could just write that.  We should also say that the distance field is 0 on the outside of the surface, which they have in their model.}
%For the colliding triangles $f_r$ and $f_i$, $v_j$ and $v_k$ denote the corresponding vertices whereas $n_j$ and $n_k$ are the normal vectors. $\psi_{f_r}$ and $\psi_{f_i}$ compute the local 3D distance field defined by the circumcenter of each triangle as described in \cite{Tzionas:IJCV:2016}. 
% \jimei{Better to explain clearly what $\psi_{f_r}$ and $\psi_{f_i}$ stand for to make the paper self-contained. A visualization would be helpful too.} \aaron{I edited it a bit, hopefully this is clearer. I'd say that including more details is low-priority.}

The weight factors $w_{r,i} = w_{i,r}$ are determined as follows.  We observed in initial tests that setting uniform weight created unnatural motions, because of slight interpenetrations near joints such as the underarms or crotch. Such interpenetrations are not objectionable, whereas penetrations between different part of the character are. Hence, we set the weights based on geodesic distance, with a small weight for  triangles that are close together on the surface: % are geodesically close to each other. Formally, we define $w_{r,i}$ as:
\begin{equation}
    w_{r,i} = \eta^{-1} \min_{v_j \in f_r, v_k \in f_i} D(v_j,v_k)
\end{equation}
where $D(v_j,v_k)$ is the geodesic distance from vertex $v_j$ to vertex $v_k$, and the normalization $\eta$ is the maximum geodesic distance between all pairs of vertices in the mesh: $\eta = \max_{a,b} D(v_a,v_b)$. 
% \aaron{Is this correct?}
% \ruben{Yes. We normalize every vertex pair geodesic distance by the max distance so far away vertices are weighted with values that approach 1, closer vertices are weighed by values that approach 0}
%Our geodesic weights give a weight that approaches $1$ to vertices that are far away from each other, and a weight that approaches $0$ to vertices that are closer together. Next, we formulate our self-penetration energy function as follows:

\paragraph{Foot contact preservation.}
This step takes as input the foot contacts with the ground detected in the source motion as described in Section \ref{sec:contact}. The energy term preserves these contacts by minimizing velocity and height for vertices that should be in contact:
\begin{equation}
E_{\mathit{foot}} = \sum_{j \in \mathcal{C}} \frac{1}{h^B} \left(\|\dot{g}^{B}_j\|^2 + \| \left(g^{B}_j\right)^y \|^2 \right),
\end{equation}
where $g^{B}$ indicates the global joint position and $\mathcal{C}$ is the set of joint indices (i.e., heels and toes) in contact with the ground at the current time-step $t$. The first term inside the parenthesis minimizes the foot joint velocities and the second term minimizes the $y$ coordinates of the foot joint positions to be at ground level during a detected contact. 

In practice, the heel height from the ground depends on how the character was built. Heel may be at ankle height in some characters, and so, we get the heel height from reference pose, i.e., T-pose skeleton, and add it to the heel coordinates so that ground height of zero corresponds to contact.

\subsubsection{Skeleton-level motion transfer term} \label{sec:motion_modeling}

% \aaron{Is anything in this section novel? If not, acknowledge that it's standard}
% \ruben{We point this out at the beginning of the section "Like previous work, we also include terms to preserve motion....."}
%\aaron{I propose moving the ground contact terms to the previous section, so that they're all in a contact section.}
%\jimei{I agree with Aaron.}

% \aaron{is there anything novel in this section? if not, perhaps say that it's similar to previous methods, maybe give citations.}
In order to preserve the style of the input motion, we adopt an energy term that models the motion at the skeleton level similar to previous works. Specifically, we focus on preserving the local motion represented as joint rotations, the global motion represented as the root trajectory and the global motion of the end-effectors (i.e., hands and feet). Hence, our motion energy function is defined as:
\begin{equation}
    E_{\mathit{skel}} = E_{\mathit{weak}} + \omega E_{\mathit{ee}},
\end{equation}
where $E_{weak}$ models local and global motion and $E_{ee}$ models the end-effector positions. Our first term encourages the retargeted joint rotations to remain close to the source motion in a weakly supervised manner, while also encouraging the target character to follow a global trajectory similar to the source. Our weakly supervised energy is defined by:
\begin{equation}
    E_{\mathit{weak}} = \rho \sum_j\|\theta^B_j - \theta^A_j\|^2 + \|o^B - o^{A \rightarrow B}\|^2, \label{eq:weak}
\end{equation}
% \jimei{Better to add joint index}
where $\theta^A_j$ is the rotation of joint $j$ in the source motion,  $\theta^B_j$ is the rotation of joint $j$ in the retargeted motion, $o^B$ is the retargeted root velocity, and $\rho$ is the weight of this term. The 
source motion's root velocity $o^A$ is scaled by the height ratio between the legs of the source and target characters, producing $o^{A \rightarrow B}$.
% \aaron{can we write it as $s o^A$, and say $s$ is the height ratio?}
% \ruben{One problem with that is we use $s$ for skeleton in the RNN formulation}

We further define our second term which models the motion of the end-effectors as follows:
\begin{equation}
E_{\mathit{ee}} = \sum_{j \in ee}\|\frac{1}{h^B} \dot{g}^{B}_j - \frac{1}{h^A} \dot{ g}^{A}_j\|^2,
\end{equation}
where $\dot{g}^{A}_j$ and $\dot{g}^{B}_j$ respectively are the velocities of the source and output character joint $j$ in global coordinates and $ee$ is the set of joint indices of the hands and feet end-effectors.
% \jimei{like all the joints or just the root?} \ruben{all joints, but we only consider the end-effectors in this energy}
$E_{\mathit{ee}}$ minimizes the difference between the global velocities of end-effectors, scaled by the respective character heights $h^A$ and $h^B$.
% \jimei{which joints are the end effectors? Be specific.}
%
%the global joint coordinates at frame $t$, $\frac{1}{h^B}\left(g_t^{B} - g_{t-1}^{B}\right)_{\text{ee}}$ and $\frac{1}{h^A}\left(g_{t}^{A} -g_{t-1}^{A}\right)_{\text{ee}}$ models the global velocities of the end-effectors of the retargetted and source skeletal motion at frame $t$ scaled by their respective character height.

%Finally, the third term preserves foot contact with the ground, by minimizing velocity and height for vertices that should be in contact:
%\begin{equation}
%E_{\mathit{foot}} = \sum_{j \in \mathcal{C}} \frac{1}{h^B} \left(\|\dot{g}^{B}_j\|^2 %+ \| \left(g^{B}_j\right)^y \|^2 \right),
%\end{equation}
%where $g^{B}$ indicates the global joint position and $\mathcal{C}$ is the set of joint indices (i.e., heels and toes) in contact with the ground at the current time-step $t$, as determined in Section \ref{sec:contact}. The first term inside the parenthesis minimizes the foot joint velocities and the second term minimizes the $y$ coordinates of the foot joint positions to be at ground level during a detected contact. 
%
%In practice, the heel height from the ground depends on how the character was built. Heel may be at ankle height in some characters, and so, we get the heel height from reference pose, i.e., T-pose skeleton, and add it to the heel coordinates so that ground height of zero corresponds to contact.

% \ruben{mention we do autoencoder training also and try to regress to the joints too similar to NKN}

\subsection{Geometry-Conditioned RNN} \label{sec:geocond_rnn}
Directly optimizing the output motion is very expensive and impractical, which has motivated the development of RNN-based methods \cite{fragkiadaki2015recurrent}. Direct optimization would be especially impractical for our case, since geometric collision detection on full-body meshes is memory-intensive.
% \jimei{We have all the terms in the energy function. I think it is possible to optimize it directly without a RNN  like what the training step does. Will it be too expensive or too inefficient? Ruben can confirm.}
% \ruben{It is too expensive to fit skinned motion at all frames if we consider the full sequence and the sequence is long (hundreds / thousands of frames), and if I try to optimize one frame at a time as in our method, the optimization fails. I experienced this even while trying to do Jacobian IK for foot contacts which should be easier (thank God our encoder-space optimization worked). It requires to provide the optimized trajectories to perform IK which is hyper-parameter dependent if smoothing needs to be done at the contact points, and so, it's very sensitive to the accuracy of contacts. I probably also needed to optimize contact timing an other things that I didn't do. During my tries, I had to isolate the legs from the rest of the upper-body to not destroy the upper-body joints too much, and even then, some motions had a lot of artifacts. It basically requires too much intervention to make it work properly, and possibly domain knowledge that I likely dont have.}

Instead, we train a recurrent neural network (RNN) that retargets a source motion to a target character by minimizing the energy function defined in the previous section. As shown in Figure~\ref{fig:network}, the RNN is composed of an encoder that encodes the source skeleton motion $m^A$ into RNN states $h^{\textit{enc}}$ and a decoder that outputs motion features in the hidden states that are used to predict the target skeleton motion, $m^B$, and skinned motion, $\hat{v}^B$. In order to decode hidden motion features, the decoder takes as input the encoded source motion features in the current frame, the local motion ($p^B$) and root velocity ($o^B$) of the target character in the previous frame as well as the skeleton ($\bar{s}^B$) and the geometric encoding of the target character in reference pose (i.e., T-pose). We utilize the PointNet \cite{pointnet} architecture to obtain a geometric encoding of the target character given its geometry and skinning weights in the reference pose. This enables our architecture to be invariant to the topology of the target character mesh. Given the hidden motion features, we utilize linear layers to decode the local joint rotations and the root velocity. A Forward kinematics (FK) layer converts the local joint rotations ($\theta^B$) to local joint positions and by adding the root velocity we obtain the global skeleton joint coordinates, $g^B$. We also incorporate a skinning layer (SK) that deforms the target character geometry based on the skeleton motion and outputs the local skinned motion, $\hat{v}^B$. We refer the reader to the supplementary material for details on the network architecture.

\input{figures/network}

% \aaron{I suggest we put the figure back if there's space.}

\subsection{Encoder-Space Optimization} \label{sec:in-net-optim} 

While the RNN produces good output motions, they often do not adequatenly satisfy contacts and penetration constraints.  
%\aaron{can we be more precise about how the RNN result is inadquately perceptually? e.g., old text was "some contacts may cause significant self-penetrations in certain characters so the learned model will try to avoid them"} 
%We can refine the RNN result by optimizing the output motion further. However, spacetime optimization of the full motion remains impractical. 
Therefore, at test-time, we take the RNN output as the initial solution to our energy function and continue optimizing it. As directly optimizing the pose representation remains impractical, we propose \textit{encoder-space optimization}, in which the motion is updated by updating the hidden encoding $h^{\mathit{enc}}$ and root velocity $o$.  This allows us to update the motion frame-by-frame, while also taking advantage of the smooth, low-dimensional, decoupled embedding learned by the network. 
%\input{figures/innet_optimization}
%The proposed neural network architecture in Section \ref{sec:geocond_rnn} can learn accurate contact-aware constraints directly from data. However, some contacts may cause significant self-penetrations in certain characters so the learned model will try to avoid them. A simple solution would be to perform a post optimization step using our proposed energy function in Equation \ref{eq:full}. Nevertheless, such numerical optimization methods can be expensive and difficult to apply. Therefore, we propose to use an in-network optimization step which optimizes the hidden encoded motion features, $h_{\text{enc}}$, as well as the root velocity, $o$, in our neural network such that difficult contacts are achieved (Figure \ref{fig:innet_optim}). Namely, our in-network optimization is performed by the following update rules:
Specifically, for each time-step $t$ of the output motion, we perform $N=30$ gradient descent updates to the encoder hidden units and root velocity as follows:
\begin{align}
h_{t,n+1}^{\text{enc}} & \leftarrow h_{t,n}^{\text{enc}} - \alpha \frac{\partial E_{\mathit{full}}(t)}{\partial h_{t,n}^{\text{enc}}}, \\
o_{t-1,n+1} & \leftarrow o_{t-1,n} - \alpha \frac{\partial E_{\mathit{full}}(t)}{\partial o_{t-1,n}},
\end{align}
where $1\le n \le N$ indicates the iteration number. Note that these updates only consider the $E(t)$ terms that depend on the current time-step $t$. The output motion can then be generated from the updated hidden units and root velocity.

Attempting to sequentially update the pose parameters $\theta$ frame-by-frame does not improve the motion, since these parameters are tightly coupled in the energy.
%where the subscript $n$ indicates the current in-network optimization iteration. We run our in-network optimization for $N$ steps for each frame $t$ (We use $N=10$ in our experiments). After every update, we feed the updated $h_{t,n}^{enc}$ to our RNN decoder. Optimizing for our energy function in the hidden unit space allows us to take advantage of 1) the already good enough solution that our network outputs at $n=0$, and 2) the motion space the neural network has learned during training which simplifies the optimization. In contrast, optimizing in joint rotation space proved to be complex and leads to instabilities during the optimization (Section \ref{sec:comparisons}). \ruben{update figure with in-net optimization of the hip offset} \ruben{mention that updating for the full energy is easier when the network has been trained with it}

\subsection{Input contact detection} \label{sec:contact}
% Collisions between meshes are detected by efficiently
% finding the set of colliding trianges C using bounding
% volume hierarchies (BVH)

%One of the main components of our retargeting energy is the ability to preserve meaningful contacts of the character with itself that play a key role in terms of reflecting the characteristics of the source motion. Hands placed together to convey a begging motion or a hand touching the head to perform a particular dance move are examples of such contacts. We observe that hands are often involved in these contacts so without loss of generality we focus our contact detection algorithm to hands. 

We now describe how we estimate self-contacts and ground contact in the source motion. 

\paragraph{Self-contacts.}
There are two steps to our self-contact detection process. First, we identify instances in the source motion where either of the character's hands intersect any other body part, including the other hand. Then, we convert these intersections into contact constraints for use in the output motion.

As a preprocess, we group the vertices associated with skinning weights over 0.5 for each joint. For example, the left hand group is the set of vertices for which the skinning weight from the left wrist joint is at least 0.5. Each group also includes the triangles formed by its vertices.

In a given input frame, the algorithm detects if there is an intersection between one of the hand groups and another group on the body, using a Bounding Volume Hierarchy (BVH) \cite{egst.20041028} on the triangles in the groups.  The two groups are determined to be in contact if the average cosine similarity of the per-vertex velocities in global coordinates is greater than $0.9$, or if the distance between their nearest vertices is less than $0.2~\mathrm{cm}$ where the shortest character in our dataset is $138~\mathrm{cm}$.
%For a given input frame, we first detect intersections between hand triangles, and any triangle elsewhere on the other hand or the rest of the body, using a Bounding Volume Hierarchy (BVH) \cite{egst.20041028}.
%We identify the hand vertices of the input model as those that have a skinning weight of at least 0.5 on either hand joint.
%If an intersection is found, we identify the other joint involved in the intersection, as well as the set of vertices that belong to this joint by a similar skinning weight threshold.
% \aaron{what if the intersection involves more than one joint? surely they all do. need to introduce the idea of groups of vertices slightly earlier, there's nothing about a hand group of vertices here.}
% \ruben{If hands intersect two groups of vertices, we register vertices from the two groups during the self-contact optimization and average them}
%Given the two groups of vertices, we mark them as in contact if the average cosine similarity of their velocity is greater than $0.9$ or if the minimum distance between them is less than $0.2~\mathrm{cm}$ in a $180~\mathrm{cm}$ scale.
% (\duygu{I'm not sure how we have the units here?}).
For each detected contact, we identify the top 3 closest pairs of vertices between the two groups.
The same process is repeated for all pairs of intersecting groups.
%Note that if intersections involve more than one group, we use all intersection groups and minimize the average between the vertices found to be in contact in each group.
% \jimei{needs a figure with a zoom-in look at hand mesh to visualize the self-contacts and how they are transferred to target mesh. Hand vertices and their corresponding contact vertices can be color-coded differently. If time doesn't allow it, we could put it in the supp.} \ruben{Let's try to put this in the supplementary material. I have a visualization, but it's matplotlib which is not professional looking}

Since the input and output characters may have very different mesh geometry and topology, we next need to transfer contact constraints to the output mesh.  Let $v^A$ be a contact vertex in the input shape; we first must identify its corresponding vertex $v^B$ in the output shape. As in many previous works, we use feature matching for mesh correspondence. Specifically, we define a per-vertex feature vector that combines skinning weights and offset vectors from the corresponding joint positions in the reference pose; we found that skinning weights alone were inadequate for correspondence.  Then, the corresponding output mesh vertex $v^B$ is the output vertex with the most-similar feature vector to the input vertex~\cite{wei_2016}.  Given a pair of input vertices in contact $v^A_i$ and $v^A_j$ on the input, we generate a constraint that the corresponding vertices $v^B_i$ and $v^B_j$ should be in contact on the target mesh. Details of the feature vector are in the supplemental material. 
%\aaron{cite other works that perform mesh correspondence based on nearest feature vectors?}
%\aaron{this is a place where we could communicate the difficult/decisions that went into this? it wasn't that easy to formulate, as I recall? or am I thinking of the interpenetration energy...}
%\ruben{If I remember correctly, we tried different feature vectors. First, we tried to simply use skinning weights. Then we tried skinning weights + offset and stuck with it}

%In a general setting, the source and target characters are expected to have skin geometries with arbitrary topology and number of vertices. Therefore, given the set of source vertex pairs involved in each contact, we first find the corresponding vertices in the target character. We do this by first computing a per-vertex feature vector combining the skinning weights and the offset vectors from the corresponding joint positions in the reference pose. We then simply compute closest point correspondences in the feature space. Once we find the corresponding set of target vertex pairs involved in each contact, we encourage these vertices to stay close to each other:

\paragraph{Foot contacts.} We find that simple thresholds similar to previous work, e.g., \cite{rempe2020contact} are sufficient for detecting foot contacts with the ground. The feet are defined by the toe and heel joints, which we threshold in different ways. The toe joint is determined to be in contact if its height from the ground is at most 3cm and displacement from previous time-step is at most 1cm, all at 180 cm scale. The heel is determined to be in contact only if its displacement from the previous time-step is at most 1 cm at 180 cm scale. We treat the heel differently from the toe because some of the motions were captured with subjects wearing high heel shoes which makes our threshold based on heights from the ground inaccurate. We perform an additional step to remove any false negatives by a sliding window approach that fills in contacts if more than half of the frames in a window of 5 frames are in contact.
% \aaron{tbd.}

%% file: figures/overview.tex
\begin{figure*}[htp!]
  \centering
  \includegraphics[width=.88\linewidth]{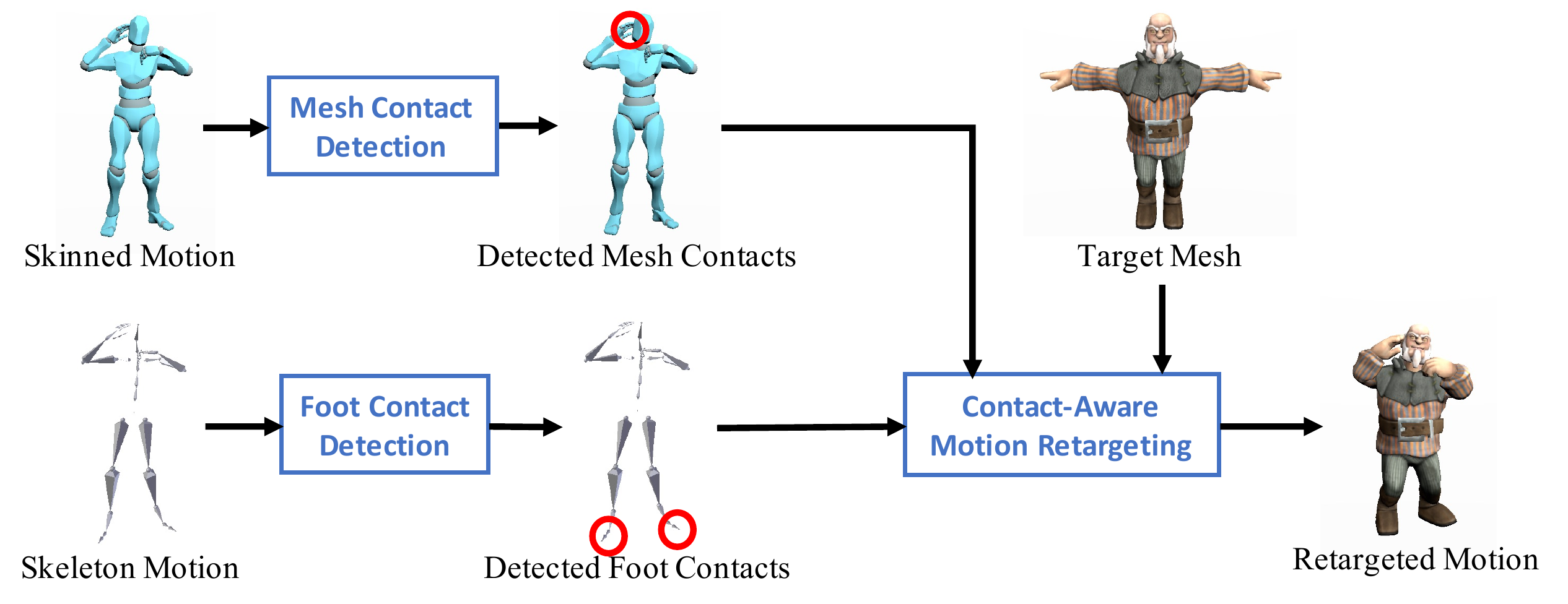}
  \vspace{-.1in}
  \caption{Method overview. Our method first detects hand contacts in the input motion geometry and foot contacts with the floor. The detected contacts are then passed into our contact-aware motion retargeting which retargets the source motion into the target character while preserving the detected contacts.}
  \label{fig:overview}
  \vspace{-.2in}
\end{figure*}

%% file: figures/network.tex
\begin{figure}[t!]
  \centering
  \includegraphics[width=0.95\linewidth]{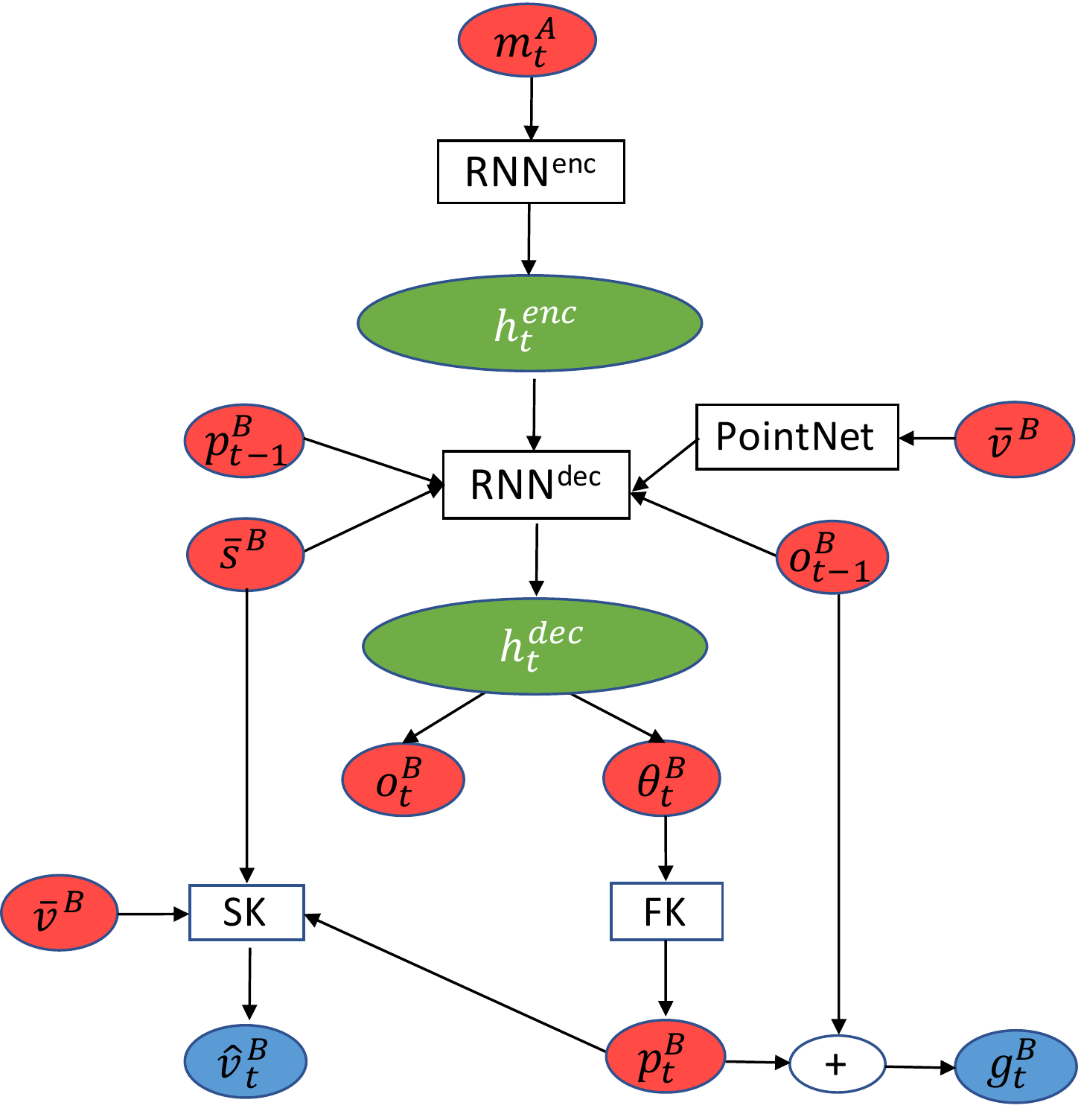}
  \caption{Geometry-conditioned RNN. See text for details.}
  \label{fig:network}
  \vspace{-.1in}
\end{figure}

% \begin{figure*}[t!]
%   \centering
%   \includegraphics[width=1\linewidth]{figures/network4.pdf}
%   \caption{Geometry-conditioned RNN.}
%   \label{fig:network}
% \end{figure*}

%% file: sections/4_experiments.tex
\section{Experiments}
\paragraph{\textbf{Dataset, training and testing.}}
We evaluate our method on the Mixamo dataset~\cite{Mixamo}, which consists of sequences of the same motions performed by multiple characters with different body shapes and skeleton geometries. 
 
A shortcoming of the Mixamo dataset is that it does not provide clean ground truth: many of these motions exhibit the kind of interpenetration and ground contact violations that we aim to correct, and no perfectly clean dataset is available. These issues may not be a problem during training because contacts are enforced by our loss terms rather than being purely learned. However, evaluation is difficult for these motions; it is very difficult to visually evaluate the quality of motion transfer when the input suffers from contact and penetration issues. Hence, for all of our tests, we use the ``Y-bot'' character motions as source motion, because these motions do have accurate self-contacts.
% ~\jimei{I'm not sure if we should mention that we learned from mixamo team that Y-box motion are the original motion so they are more accurate, or we just say "we observe Y-bot motion are more accurate"}.
The ``Y-bot'' character is not seen during training.

We train on the training subset used in \cite{Villegas_2018_CVPR}, which contains distinct motion sequences performed by 7 characters, and contains 1646 motions in total.
Similar to \cite{Villegas_2018_CVPR,Lim_pmnet_BMVC19}, we train our method by extracting clips of 60 frames long, which we retarget to characters in the training.
In contrast to \cite{Villegas_2018_CVPR,Lim_pmnet_BMVC19}, we evaluate our method and baselines on full motion capture sequences of variable length.
Please see the supplementary materials for training and testing details.

\paragraph{\textbf{Preprocessing.}}
We preprocess each motion sequence similar to \cite{Villegas_2018_CVPR,Lim_pmnet_BMVC19,Holden2016} by separating into local pose and global pose. While local pose consists of joint rotations relative to their parents, global pose consists of the velocity of the root (i.e., hip) in $x$, $y$, and $z$ coordinates. We use the same 22 joint subset modeled in \cite{Villegas_2018_CVPR,Lim_pmnet_BMVC19}. Please see the supplementary materials for more details.
% Similar to \cite{Villegas_2018_CVPR,Lim_pmnet_BMVC19}, we use a subset of 22 joints from the common Mixamo skeleton, ignoring the fingers and toes. Please see the supplementary material for the details. \todo{put joint names list in supp}

\input{tables/comparisons}

\paragraph{\textbf{Baselines.}}
We compare against the unsupervised motion retargeting methods from \cite{Villegas_2018_CVPR} and \cite{Lim_pmnet_BMVC19} which both involve a forward kinematics layer in their network architectures. We also compare with Skeleton-Aware Networks \cite{aberman2020skeleton} (SAN) which uses skeleton pooling-based encoder and decoder networks to learn a skeleton invariant representation via an adversarial cycle consistency method. Note that SAN is an offline motion retargeting method (i.e., it requires the whole source motion sequence to be seen before retargeting), and applies inverse-kinematics (IK) as post-processing to maintain foot contacts in the final output. Following SAN, we also include a baseline where we take our network outputs and perform the same IK as a post-processing step as in \cite{aberman2020skeleton} instead of our encoder-space optimization.
% Following SAN, we experimented with their IK post-processing step, but found that it did not completely remove foot skating in comparison to our encoder-space optimization.
% \aaron{we need to discuss then how our method works if you run the RNN and then apply IK as a post-process.}
% \ruben{mention that SAN is an offline retargeting method (easier to do), and they have to do IK as post-optimization to fix foot contacts}

% For our optimization based baselines, we compare against the industry popular Autodesk HumanIK algorithm. 
% We use two types of baselines: Data driven and optimization based. For our data driven baselines, we compare against the unsupervised motion retargeting methods from \cite{Villegas_2018_CVPR} and \cite{Lim_pmnet_BMVC19} which both involve a forward kinematics layer in their network architectures. We also compare against the recently proposed skeleton-aware network \cite{aberman2020skeleton} which uses an edge collapse based encoder and decoder networks to learn a skeleton invariant representation via an adversarial cycle consistency method.
% For our optimization based baselines, we compare against the industry popular Autodesk HumanIK algorithm. 
%Finally, we also use our energy function to optimize the HumanIK retargetted motion and show that simply using our energy function to optimize in joint rotation space rather than learned feature space does not improve the performance. \ruben{Not sure what to say here. I was not able to get anything reasonable by doing gradient descent on the angles using our objective on the Human IK outputs}

\paragraph{\textbf{Evaluation metrics.}}
We evaluate results based both on the geometry and the skeleton.
We evaluate skeletal motion, by measuring how close the retargeted motions are to the ground-truth provided in the Mixamo dataset. 
The global joint position error is normalized by character height. 

Interpenetration is scored using the interpenetration penalty in Equation~\ref{eq:int} on the normalized vertices. 
% \jimei{vertices will lie in unit sphere once normalized, so no need to mention "unit sphere" unless it means something else.} \ruben{I think you can also normalize them within a cube, so I wanted to make sure it's clear we do sphere and not cube. Do you mean this is already mentioned somewhere else?}
We evaluate how well self-contacts are preserved by measuring the distance between the vertices involved in each contact. We use continuous scores rather than binary losses, because some characters are unable to perfectly reach a contact point due to their geometries (e.g., bulky vs skinny character), and slight contact errors are preferable to large ones. 
% \footnote{Note that the ground-truth data in Mixamo is generated using the Autodesk HumanIK motion retargeting software. \jimei{again, shall we mention we obtain this information from mixamo team?}}
%in terms of global joint positions, and how well foot contacts are preserved. 
Our foot contact metric is based on a binary classification of the foot contacts detected in the retargeted motion against the contacts detected in the source motion. We use a binary metric since, in contrast to our vertex distance metric, there is nothing preventing feet from reaching the ground. 

Finally, we conduct a user evaluation where we show two retargeting results of a given source motion, our result versus the result of the best performing baseline method, and ask the users to choose the more plausible result. 

%1) How accurately the retargetted skinned motion is modeled, 2) how accurately the retargetted skeletal motion is modeled, and 3) a user study to evaluate the perceptual plausibility of the retargeting. For the first evaluation criteria, we measure the amount of self-penetrations caused by the retargetted motion using our self-penetrations penalty, and the distance between vertices that should be in contact in the retargetted motion. The reasoning for the distance based vertex contact evaluation is that there are some characters that should not be able to reach a contact point without causing large self-penetrations. However, the retargetted motion should at least get the vertices that should be in contact as close as possible to each other. For the second evaluation criteria, we measure how close the retargetted skeletal motion is to the ground-truth motion in Mixamo \footnote{Note that the ground-truth data in Mixamo is generated using HumanIK which is one of our baselines.}. Namely, we evaluate the local joint locations with respect to the hip joint, the joint coordinates in world space, and how well foot contacts are modeled. For the third evaluation, we provide users a video of the input motion and two retargetted motions. We let them decide which of the retargetted motions look the closest to the input motion.

\subsection{Results}
In this section, we present quantitative and qualitative experiments.
Please see our supplementary material for an ablation of the different components in our method.

\input{figures/qualitative}

\subsubsection{Comparisons} \label{sec:comparisons}
In this section, we compare our full method against the recently proposed motion retargeting methods. As shown in Table \ref{table:comparison}, our method outperforms all the baselines not only in terms of self-contact handling and interpenetration, the focus of our method, but also in terms of motion quality. A significant advantage of our method is the encoder-space optimization to enforce constraints. In contrast, non of the baseline methods explicitly model the contacts. The SAN method \cite{aberman2020skeleton} presents an additional post-processing step based on Inverse Kinematics (IK) to ensure foot contacts are modeled (SAN + IK). While it does improve foot contact accuracy and global position MSE, it sacrifices the geometry quality. 
%however, we found that this post-processing optimization is not as effective as our encoder-space optimization as shown in Table~\ref{table:ablation}.
To have a fair comparison, we perform the same IK on the outputs of our network without encoder-space optimization ($E_{\mathit{full}}$ + IK). It is clear that even though foot contact accuracy is improved with IK, our complete method ($E_{\mathit{full}}$ + ESO) outperforms it by a large margin. We invite readers to watch the \href{https://www.youtube.com/watch?v=qQ4HO2Hibsk}{supplementary videos} for more comparisons and animated motions.

\paragraph{User study.}
\input{tables/user_study}
We conduct a user study to qualitatively evaluate the performance of our method against SAN + IK, the best performing baseline. For each question, human subjects are given three videos, namely the source motion, and the two motions retargeted using either our method or SAN + IK. The retargeted results are randomly placed on either side of the source motion and anonymously labeled. We ask the subjects to choose the retargeting result that looks closest to the source motion.
Each subject is shown 20 questions where the motion triplets are randomly sampled out of 180 test motion sequences. 
We run our user study on a total of 17 subjects where 8 subjects have animation expertise (animation artist or developer) and 9 are not, collecting a total of 340 comparisons. Further details of how we conduct the user evaluation is provided in the supplemental material.

Our results are summarized in Table \ref{table:user_study}. We find that overall 80\% of users prefer our retargeted motion over the baseline. Furthermore 86\% of the expert subjects prefer our results demonstrating the superior performance of our approach. We also provide a qualitative comparison of our method against the baseline in Figure \ref{fig:qualitative} where we highlight the penetration reduction and self-contact capabilities of our method in comparison to the baseline.

%\aaron{study details should be in the supplemental material.}

\subsection{Retargeting motion from human videos}
In this section, we test our method for retargeting motion estimates from human videos.
We get the human pose estimates from \cite{rempe2020contact}, and feed it to our method to retarget motion that takes the character mesh into account.
In Figure~\ref{fig:qualitative_humans1}, we show how our method (bottom row) is able to avoid interpenetration present in the motion estimated by \cite{rempe2020contact}.
This happens because their method processes character skeleton while completely ignoring the character mesh.
In addition, we test our method for retargeting human motion into characters with a significant body shape difference.
In Figure~\ref{fig:qualitative_humans2}, we show how the motion retargeted using our method follows the input video closely while moving within the target character body shape constraints.
Please refer to our \href{https://www.youtube.com/watch?v=qQ4HO2Hibsk}{supplementary material} for video results.

\input{figures/qualitative_humans1}
\input{figures/qualitative_humans2}

%% file: tables/comparisons.tex
% \begin{table*}%[htp!] 
% % \small
% \centering
% \begin{tabular}{l||c|c||c|c}
% \Xhline{4\arrayrulewidth}
% \multirow{3}{*}{ Method }  & \multicolumn{2}{c||}{Geometry evaluation} & \multicolumn{2}{c}{Motion evaluation} \tabularnewline
% \cline{2-5} 
%  & \makecell{ Inter-Penetrations $\downarrow$ } & \makecell{Contact Distance}  $\downarrow$& \makecell{Foot Contacts \\ (Prec / Recall) $\uparrow$} & \makecell{Global Pos. \\ (MSE) $\downarrow$} \\
% \Xhline{4\arrayrulewidth}
% Ours & \textbf{4.70} & \textbf{3.87} & \textbf{0.98 / 0.97} & \textbf{0.48} \\
% SAN \cite{aberman2020skeleton} & 5.95 & 5.26 & 0.94 / 0.45 & 0.82  \\
% SAN \cite{aberman2020skeleton} + IK & 7.40 & 8.42 & 0.96 / 0.62 & 8.13  \\
% PMnet & 13.46 & 23.11 & 0.97 / 0.54 & 3.67  \\
% NKN & 10.98 & 14.86 & 0.95 / 0.61 & 8.15  \\
% % HumanIK* & 6.59 & 5.11 & \textbf{0.98 / 0.97} & - \\
% % HumanIK + Our Energy & - & - & - & - \\
% \Xhline{4\arrayrulewidth}
% \end{tabular}
% % \vspace{-.1in}
% \caption{Motion retargetting evaluation. We evaluate the retargetted motion at the geometry and skeletal level. We evaluate the amount of self-penetrations, geometry contacts distance, foot contacts with the floor, and global joint positions in the retargetted motion. }
% % \vspace{-0.1in}
% \label{table:comparison}
% \end{table*}

\begin{table*}%[htp!] 
% \small
\centering
\begin{tabular}{l||c|c||c|c}
\Xhline{4\arrayrulewidth}
\multirow{3}{*}{ Method }  & \multicolumn{2}{c||}{Geometry evaluation} & \multicolumn{2}{c}{Motion evaluation} \tabularnewline
\cline{2-5} 
 & \makecell{ Inter-Penetrations} $\downarrow$ & \makecell{Vertex Contact \\ MSE}  $\downarrow$& \makecell{Foot Contact \\ Accuracy} $\uparrow$ & \makecell{Global Position \\ MSE} $\downarrow$\\
\Xhline{4\arrayrulewidth}
Ours & \textbf{0.81} & \textbf{3.87} & \textbf{0.97} & \textbf{0.48} \\
$E_{\mathit{full}} $ + IK & 1.19 & 4.52 & 0.82 & 1.58 \\
SAN \cite{aberman2020skeleton} & 1.43 & 5.26 & 0.63 & 0.82  \\
SAN + IK \cite{aberman2020skeleton} & 1.32 & 5.78 & 0.73 & 0.74  \\
PMnet \cite{Lim_pmnet_BMVC19} & 2.94 & 23.11 & 0.70 & 3.67  \\
NKN \cite{Villegas_2018_CVPR} & 3.20 & 14.86 & 0.71 & 8.15  \\
% HumanIK* & 6.59 & 5.11 & \textbf{0.98 / 0.97} & - \\
% HumanIK + Our Energy & - & - & - & - \\
\Xhline{4\arrayrulewidth}
\end{tabular}
% \vspace{-.1in}
\caption{We evaluate the retargeted motion at the geometry and skeletal level. We evaluate the amount of self-penetrations, geometry contacts distance, foot contacts with the floor, and global joint positions in the retargeted motion.}
% \vspace{-0.1in}
\label{table:comparison}
\end{table*}

%% file: figures/qualitative.tex
\begin{figure*}[t!]
  \centering
  \includegraphics[width=1\linewidth]{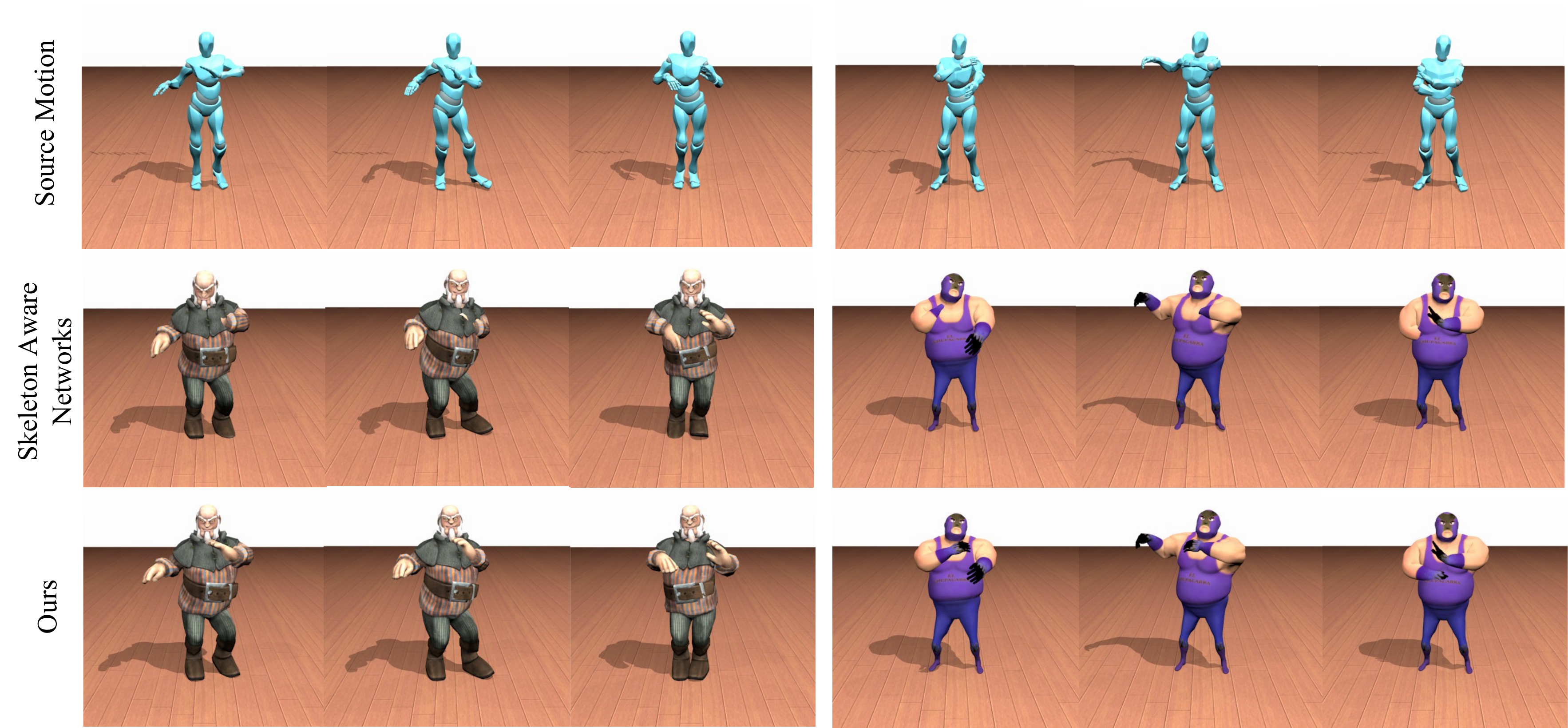}
  \caption{Qualitative comparison. We show an example where our method is able to reduce interpenetration without the need of contact modeling (left side), and an example where our method is able to model self-contacts while reducing self-penetrations (right side). We invite readers to watch the \href{https://www.youtube.com/watch?v=qQ4HO2Hibsk}{supplementary videos} for a more detailed visual comparison.}
  \label{fig:qualitative}
%   \vspace{-.1in}
\end{figure*}

%% file: tables/user_study.tex
\begin{table}%[htp!] 
\small
% \vspace{.1in}
\centering
\begin{tabular}{c|c|c}
\Xhline{4\arrayrulewidth}
\makecell{Expert User \\ Ours / SAN + IK} & \makecell{Non-Expert User \\ Ours / SAN + IK} & \makecell{Overall \\ Ours / SAN + IK} \\
\Xhline{4\arrayrulewidth}
\textbf{0.86} / 0.14 & \textbf{0.74} / 0.26 & \textbf{0.80} / 0.20 \\
\Xhline{4\arrayrulewidth}
\end{tabular}
% \vspace{-.1in}
% \caption{User study. We provide video comparisons of the two best performing methods to expert and non-expert animators. We ask them to choose the retargetted motion that is closer to the source motion and show the percentage of times each method is chosen.}
\caption{We ask human subjects to compare our retargeting results to that of the best performing baseline in Table \ref{table:comparison}. Overall $80\%$ of the users prefer our results.}
\vspace{-0.1in}
\label{table:user_study}
\end{table}

%% file: figures/qualitative_humans1.tex
\begin{figure}[t!]
  \centering
  \includegraphics[width=1\linewidth]{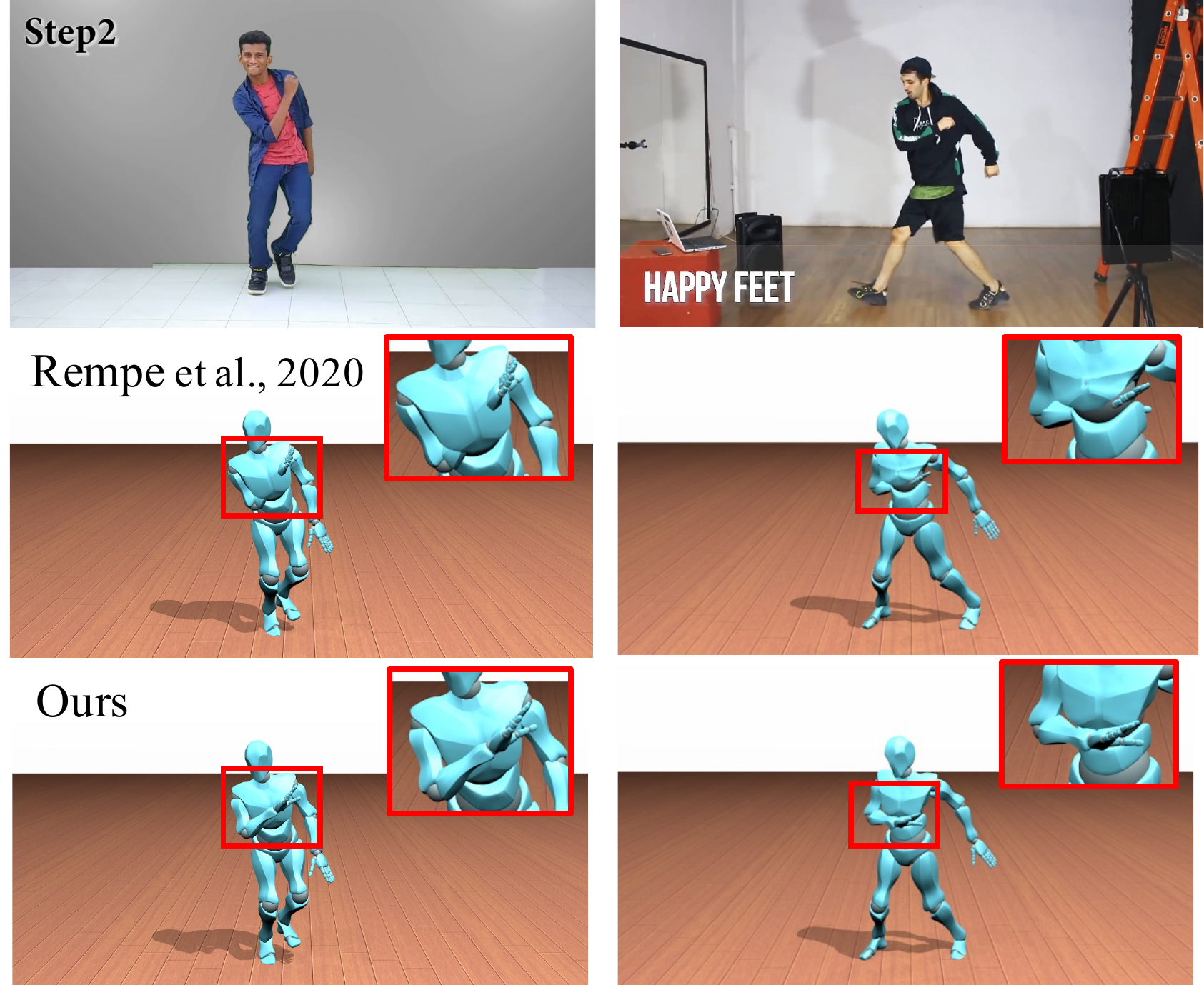}
  \caption{We retarget two motion clips estimated using the method by \cite{rempe2020contact}, and synthesize motion that avoids interpenetration present in the original input. We invite readers to watch the \href{https://www.youtube.com/watch?v=qQ4HO2Hibsk}{supplementary videos} for comparisons.}
  \label{fig:qualitative_humans1}
  \vspace{-.2in}
\end{figure}

%% file: figures/qualitative_humans2.tex
\begin{figure}[t!]
  \centering
  \includegraphics[width=1\linewidth]{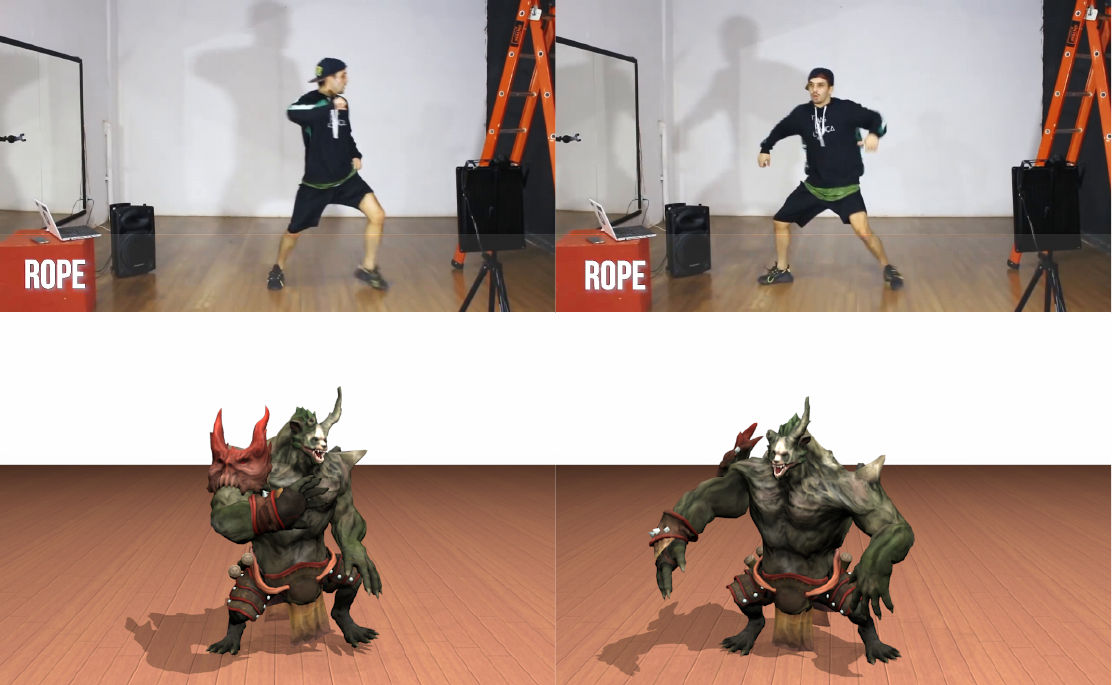}
  \caption{We retarget motion estimated by \cite{rempe2020contact} into a character with significant body shape differences. We invite readers to watch the \href{https://www.youtube.com/watch?v=qQ4HO2Hibsk}{supplementary video} for more results.}
  \label{fig:qualitative_humans2}
  \vspace{-.2in}
\end{figure}

%% file: sections/5_conclusion.tex
\section{Conclusion}
We present a motion retargeting method that detects and preserves self and ground contacts while also reducing interpenetration. At the core of our method is an energy formulation that enforces such geometric constraints while also preserving the input motion quality. We train a recurrent neural network (RNN) that minimizes this energy function to retarget motion between source and target characters that vary in terms of skeleton lengths, proportions, as well as character geometries. We further propose an encoder-space optimization strategy to refine the hidden encoding of the RNN to satisfy the contact constraints.

While our method outperforms recently proposed retargeting solutions both quantitatively and qualitatively, there still remain some limitations that we would like to address in future work. The different energy terms in our formulation quantify various qualities of a good retargeting output. However, occasionally different terms might be conflicting, e.g., the retargeting might sacrifice preserving the input motion in order to reduce interpenetration. This might result in the loss of the characteristic motion style of the input. A thorough analysis of the factors that affect the perception of a motion and how they can be incorporated in a retargeting method is an interesting future research direction. In our current implementation, we use a heuristic approach to detect the self-contacts in the input motion. Provided with annotated data, such contacts can be learned automatically. Enforcing the interpenetration term aggressively may result in stiff retargeted motion. We utilize geodesic weights to softly penalize intersections in regions such as underarms. Providing the user with the ability to paint regions that allow slight penetrations is another future direction. Finally, even though our contact-aware formulation is general, our network cannot handle arbitrary number of joints. This is an additional future direction we plan to pursue.
% This is another future direction towards a complete data-drive motion retargeting method.

%% file: sections/7_acknowledgements.tex
\section*{Acknowledgements}
We would like to thank Minhyuk Sung for his help during initial stages of the project. We would also like to thank Stefano Corazza and everyone in Adobe Aero and Adobe Research for their helpful feedback. Last but not least, we would like to thank the participants in our user study.

%% file: sections/6_supplementary.tex
\clearpage
\onecolumn
\section*{\fontsize{15}{18}\selectfont Appendix}
\begin{appendix}
% \section{Video comparisons}
% Please see our \href{https://www.youtube.com/watch?v=qQ4HO2Hibsk}{supplementary video} for video comparisons. %The video playback has been tested using VLC player and Apple's Quicktime.

\section{Ablation study}
\input{tables/ablative_study}
In order to demonstrate the effectiveness of the different terms of our energy function (Equation~\ref{eq:full}), we provide a detailed ablation study in Table \ref{table:ablation}.

We first evaluate the performance of our network trained with the skeleton-level motion modeling term, $E_{\mathit{skel}}$ which includes the local and global motion terms and the end-effector motion terms. While using only the skeleton term results in lower global position error, we observe worse foot contact modeling and interpenetration in comparison to the other baselines.

We next evaluate the different terms involved in the geometry based energy. First, we add the foot contact preservation term, $E_{\mathit{foot}}$. This version results in a significant boost in terms of handling foot contacts. Then, we include the interpenetration term $E_{\mathit{int}}$ without the geodesic based vertex weighting, $w_{r,i}$, which results in the best interpenetration reduction, but worst vertex contact MSE. This is because the retargeted motion tries to avoid all interpenetration, including those that involve nearby vertices that are often not noticeable by viewers. This turns out to result in a stiff motion where most self-contacts are avoided. Adding the weight, $w_{r,i}$, addresses this issue by relaxing the range of the character motion. We observe that slight interpenetration occurs, but the overall motion quality and contact handling improve. We then include the vertex contact modeling term, $E_{\mathit{j2j}}$, effectively using our full energy function $E_{\mathit{full}}$. This results in a slight degradation in terms of interpenetration and motion modeling, but better self-contact handling.
% \jimei{self-contact alone or both self- and ground contacts?}. %We note that the full energy function becomes harder to learn, and some terms may take over based on how difficult they are to model. However, our geometry-conditioned RNN is able to recognize signals from our full energy function, and we can optimize its learned encoder-space towards satisfying hard constraints in the output motion.
Finally, incorporating the Encoder-Space Optimization (ESO) to satisfy hard constraints achieves the best overall performance both in terms of motion quality, foot contact, and self-contact handling.

For sake of completeness, we also check whether we can achieve accurate foot contacts by simply applying the IK post-optimization step in \cite{aberman2020skeleton}, $E_{\mathit{full}} + \textit{IK post-process Optim.}$, and find that it improves on foot contacts accuracy, as easier task than full energy, but does not outperform our encoder-state optimization.

\section{Geometry-Conditioned RNN Details}
The input to the network is the source skeletal motion $m_{1:T}^A \in \mathds{R}^{267}$ and the target character represented as a set of skeleton joint offsets $\bar{s}^B \in \mathds{R}^{J \times 3}$ in reference pose (i.e., T-pose) and a skin geometry with vertices $\bar{v}^B \in \mathds{R}^{V \times 3}$. The source motion (and similarly the retargeted motion) can be decomposed into (i) $p_{1:T}^A \in \mathds{R}^{J \times 3}$, local joint coordinates  with respect to the hip (root) joint, (ii) $\theta_t^A \in \mathds{R}^{9J}$, local joint rotations with respect to each parent joint, and (iii) $o_t^A \in \mathds{R}^{3}$, the global root velocity. As illustrated in Figure \ref{fig:network}, at each frame $t$, the network retargets the motion by the following:
% \aaron{this is really hard to interpret, can we give any top-down or high-level organization to this ideas? it's currently a list of definitions of symbols and equations. why this formulation?}
\begin{align}
h_t^{\text{enc}} &= f^\text{enc}(m_t^A, h_{t-1}^{\text{enc}}),\\
h_t^{\text{dec}} &= f^\text{dec}(p^B_{t-1},o^B_{t-1}, \hat{s}^B, e^B, h_t^{\text{enc}}, h_{t-1}^{\text{dec}, }), \label{eq:optim_forward}\\
\theta_t^B &= f^{\theta}(h_t^{\text{dec}}),\\
p_t^B &= \text{FK}(\theta_t^B, \hat{s}^B),\\
\hat{v}_t^B &= \text{SK}(\theta_tf^B, p_t^B, \hat{s}^B), \\
o_t^B &= f^{o}(h_t^{\text{dec}}), \\
g_t^B &= p_t^B + o_{t-1}^B,
\end{align}
where $f^\text{enc}$ is an encoder RNN, $f^\text{dec}$ is a decoder RNN, and $f^{\theta}$ and $f^{o}$ are linear layers that output rotations and the root velocity, respectively. $\text{FK}(.,.)$ is the forward kinematics layer that reposes the skeleton based on the joint rotations and $ \text{SK}(.,.,.)$ is the skinning layer that deforms the skin geometry accordingly. $h_{t}^{\text{enc}}\in\mathds{R}^d$ is state of the encoder RNN at frame $t$, $h_{t-1}^{\text{enc}}\in\mathds{R}^d$ and $h_{t-1}^{\text{dec}}\in\mathds{R}^d$ are the states of the encoder and decoder RNNs in the previous frame. $\theta^B_{t}$ are the joint rotations retargeted into character $B$, $p^B_{t}$ and $p^B_{t-1}$ are the resulting local joint coordinates after applying forward kinematics for frame $t$ and $t-1$, and $o^B_{t}$ and $o^B_{t-1}$ are the root velocities retargeted to character $B$ for frame $t$ and $t-1$. $\hat{v}_t^B$ is the skinned retargeted motion output by the skinning layer and $g_t^B$ are the retargeted motion joints in global coordinate system. Finally, $e^B$ is an embedding computed from the target geometry by:
\begin{equation}
e^B = \max_{\bar{v}^B} \quad f^\text{vert}(\bar{v}^B, w^B)
\end{equation}
where $f^\text{vert}$ is implemented as a two layer neural network that maps the concatenated vertices $\bar{v}^B\in\mathds{R}^{V' \times 3}$ in the t-pose and the skinning weights $w^B\in\mathds{R}^{V' \times J}$ into feature vectors for each vertex which are then max pooled over all extracted vertex features following PointNet. Please note that our network architecture is independent of the choice of the vertex encoder and we use PointNet since it is independent of the underlying mesh topology.

\subsection{Architecture, training and testing details}
We train our method on a single NVIDIA Tesla V100 GPU (16GB). Our encoder $f^{\text{enc}}$ and decoder $f^{\text{dec}}$ are implemented with a two-layer GRU with 512 hidden units each. Our output functions $f^{\theta}$ and $f^{o}$ are implemented with single linear layers. $\text{FK}(.,.)$ and $\text{SK}(.,.,.)$ are implemented with standard Forward Kinematics (FK) and Linear Blend Skinning (LBS) formulations. For our encoder $f^\text{vert}$, we use a two-layer MLP with ReLU activation and 256 hidden units in each layer.
During training, we use the Adam solver with $\beta_1=0.5, \beta_2=0.999, \epsilon=1e-8$ and learning rate of $0.0001$. We train for 10 epochs with dropout out rate of $0.1$, batch size of $256$ sequences, and learning rate decay of $0.95$ per epoch. At training time, our energy function hyper-parameters are set as $\lambda=1000$, $\beta=100$, $\gamma=0.1$, $\rho=1000$, and $\omega=1000$. During training, we gradually introduce the foot contact loss $E_{\mathit{foot}}$ by multiplying $\gamma$ times a weigh factor defined as $currentepoch / totalepochs$. During the first epoch the training focuses on moving the character limbs without penalizing ground contacts. LBS is GPU memory heavy, therefore, we use a sampling strategy during training to minimize GPU usage while modeling longer motion sequences (60 frames at training time). We uniformly sample a single motion frame per motion sequence and apply interpenetration and self-contact penalties the chosen frame. By randomly sampling, our network still need to learn to model the penalties in order to minimize them for any frame in the motion. Using larger accelerators (GPU or TPU) can enable our method to apply the penalties for the same sequence length used in our experiments, and so, improve its performance. In order to easily batch multiple character meshes during training, we perform a preprocessing step where we sub-sample all our training meshes to 3000 vertices. Nevertheless, our network can read in an arbitrary number of vertices via $f^\text{vert}$ at test time. During our ESO step, we optimize $h_{t,n+1}^{\text{enc}}$ and $o_{t-1,n+1}$ by using the Adam solver with $\beta_1=0.5, \beta_2=0.999, \epsilon=1e-8$ and learning rate of $\alpha=0.01$. In addition, we relax the motion, interpenetration and self-contact penalties hyper-parameters to $\rho=10.0$, $\beta=1.0$ and $\lambda=100.0$ to allow the motion to be more dynamic. In addition, we only optimize self-contacts if the initial output from our network is within 15cm of the target contact location. We do this because the initial outputs from our network should respect the target character's shape. Therefore a large distance to a contact point means it is likely not possible for the character to reach the contact point without performing non-natural motion. Moreover, we encourage hands to move smoothly from the initial output during ESO by aggregating $0.7$ of the distance from the initial vertex position and $0.3$ of the target contact vertex position as the self-contact penalty. Finally, we only allow hands to move freely by adding a distance penalty for the joints the hands reach during contact to stay close to their original positions.
% \begin{align}
% h_{t,n+1}^{\text{enc}} & \leftarrow h_{t,n}^{\text{enc}} - \alpha \frac{\partial E_{\mathit{full}}(t)}{\partial h_{t,n}^{\text{enc}}}, \\
% o_{t-1,n+1} & \leftarrow o_{t-1,n} - \alpha \frac{\partial E_{\mathit{full}}(t)}{\partial o_{t-1,n}},
% \end{align}

\newpage
\section{Test set details}
We downloaded source 30 input motions (Y bot) of variable length where the shortest motion is $26$ frames and the longest motion is $454$ frames. In addition, we download a total of $180$ target motions for evaluation which are composed of the original $30$ motions retargeted into $6$ different characters where $3$ are skinny (Kaya, Malcolm, Liam) and $3$ are bulky (Ortiz, Peasant Man, Warrok W Kurniawan). We make sure all of our test characters contain the same motions so that we can test the behavior of each of those motions retargeted to characters with different geometries. We chose motions that have $3$ key properties such as self-contacts, potential interpenetration, and also self-contact free motions. Within those key properties we can test whether our algorithm is addressing the problems we are trying to solve involving geometry constraints and skeletal motion constraints. The test set contains the motions listed in the following table:

\input{tables/test_data}

The motions in Table \ref{table:test_data} can be downloaded by searching the exact search query text in the middle column. The search result may provide more than one motion, however, the motion of interest should be the first one. Nevertheless, make sure that the search query matches the motion description when you hover your mouse over the motion name. We also provide the number of frames in the motion for additional search information.\footnote{A direct link to download test data cannot be provided due to legal constraints.}
% \todo{} The motions in Table \ref{table:test_data} can be downloaded by first searching the exact names listed in the Motion name column. The search result will provide a number of result pages. Make sure each page is displaying 96 motions. There should be a drop down menu with options to how many motions are displayed in each page. Next, we provide the page index and the motion position within the page. We count the motion position in a left-to-right / up-to-down fashion. For example, the first motion in the second row is motion at position 5. We also provide the number of frames in the motion for additional search information since there are multiple motions with the same name.\footnote{A direct link to download test data cannot be provided due to legal constraints.}
% Paper code and test set can be found at our \href{https://sites.google.com/umich.edu/camr}{project website}.

\clearpage
\newpage
\section{Skinny versus bulky character evaluation}
In this section, we present a split of our quantitative evaluation into skinny and bulky characters. In Tables \ref{table:skinny} and \ref{table:bulky}, we note the difficulty of modeling bulky characters versus skinny characters. We can see that Interpenetration and Vertex Contact MSE roughly two times worse in the bulky characters results (Table \ref{table:bulky}). From our observations, we conjecture that our bulky characters have a harder time mimicking the source motion which in our experiments comes from a skinny character (Ybot) without violating their geometric constraints. This is observed in our Global Position MSE results which are worse than the baseline which does not model geometric constraints, SAN. In contrast, our Global Position MSE and all geometry evaluation metrics from skinny character motion retargeting are the best amongst all methods (Table \ref{table:skinny}). This happens because it is easier to mimic Ybot's motion due to the target characters also having a thin body geometry.
\input{tables/skinny_comparison}
\input{tables/bulky_comparison}

\newpage
% \section{Self-contact visualization}
% \todo{Try to get some nice visualization for this}

\section{User study details}
The SAN baseline we compare against is the full method presented in their original paper using IK as post-processing. We do this to take advantage of the slight foot sliding fix performed by their standard IK post optimization method. We independently sample $20$ sequences without replacement for each subject in our study using a uniform distribution over the $180$ available sequences. For each of the $20$ samples, we choose whether our result is placed on the left or right side of the source motion with a probability of $0.5$. Finally, we label the video on the left as \textit{A)} and the video on the right as \textit{B)}. Therefore, each of our subjects get a unique set of $20$ videos with the source motion in the middle, a video labeled as \textit{A)} on the left and a video labeled as \textit{B)} on the right. They are asked choose the video that looks closest to the source motion and provide $20$ answers where each answer is either A or B. We then get their answers, check whether they picked our motion or the baseline, and compute the numbers presented in the main text.

\section{Correspondence feature vector details}
\input{figures/correspondences}
For our vertex correspondence computation, we perform nearest neighbors search using the KDTree algorithm on a feature vector based on each vertex skinning weights and offset from parent joint. We next describe how we decided the use of that specific feature vector. We experimented with $3$ different sets of feature vectors for computing correspondences between source and target mesh vertices. We started by simply using skinning weight as feature vectors, and found that it resulted in inaccurate correspondences where large portions of the target mesh map to a similar vertex in the source mesh as see on the left column in Figure \ref{fig:correspondence}. We hypothesize that this is due to ambiguities in the skinning weights were multiple vertices are transformed with the same skinning weight values. Next, we move onto using skinning weights and offsets going from a vertex to its parent joint in the character skeleton. We find the mix of the two features doing a reasonable job as seen on the middle column in Figure \ref{fig:correspondence}. Finally, we experimented with the normalized mesh vertices and find that it negatively affects the correspondence results. We hypothesize that because the source and target meshes are different in shape, using their vertices adds noise to the correspondence computation. Therefore, in our final method, we use a feature vector constructed from each vertex skinning weights and offset from its parent joint.

\section{Data processing details}
The motion data used in our experiments consists of the following joint set: Root, Spine, Spine1, Spine2, Neck, Head, LeftUpLeg, LeftLeg, LeftFoot, LeftToeBase, RightUpLeg, RightLeg, RightFoot, RightToeBase, LeftShoulder, LeftArm, LeftForeArm, LeftHand, RightShoulder, RightArm, RightForeArm, and RightHand. These 22 joints are shared across all Mixamo characters, and represent the main motion present in Mixamo animations. Character meshes in mixamo can be made of multiple parts. Therefore, we merge all mesh parts into a single mesh for each character to facilitate computations and implementation of our method.

\end{appendix}

%% file: tables/ablative_study.tex
\begin{table*}[htp!] 
% \small
\centering
\begin{tabular}{l||c|c||c|c}
\Xhline{4\arrayrulewidth}
\multirow{3}{*}{ Method }  & \multicolumn{2}{c||}{Geometry evaluation} & \multicolumn{2}{c}{Motion evaluation} \tabularnewline
\cline{2-5} 
 & \makecell{ Inter-Penetrations } $\downarrow$ & \makecell{Vertex Contact \\ MSE}  $\downarrow$& \makecell{Foot Contact \\ Accuracy} $\uparrow$ & \makecell{Global Position \\ MSE } $\downarrow$\\
\Xhline{4\arrayrulewidth}
$E_{\mathit{skel}}$ & 1.67 & 4.23 & 0.71 & 0.56 \\
$E_{\mathit{skel}} + E_{\mathit{foot}}$ & 1.59 & 4.21 & 0.80 & 0.66 \\
$E_{\mathit{skel}} + E_{\mathit{foot}} + E_{\mathit{int} \text{ \ without \ } w_{r,i}}$ & \textbf{0.35} & 7.94 & 0.82 & 1.00  \\
$E_{\mathit{skel}} + E_{\mathit{foot}} + E_{\mathit{int} \text{ \ with \ } w_{r,i}}$ & 0.56 & 6.79 & 0.81 & 0.97  \\
$E_{\mathit{full}}$ & 1.18 & 4.52 & 0.76 & 1.54 \\
$E_{\mathit{full}} +$ ESO & 0.81 & \textbf{3.87} & \textbf{0.97} & \textbf{0.48} \\
\Xhline{4\arrayrulewidth}
\end{tabular}
% \vspace{-.1in}
\caption{We show the effects of the different energy terms in our contact-aware motion retargeting method.}
% \vspace{-0.1in}
\label{table:ablation}
\end{table*}

%% file: tables/test_data.tex
\begin{table}[h!] 
\small
\centering
\begin{tabular}{l|c|c}
\Xhline{4\arrayrulewidth}
\makecell{Motion name} & \makecell{Search query} & \makecell{Number of frames} \\
\Xhline{4\arrayrulewidth}
Agreeing  & Step Back Cautiosly Agreeing & 141 \\
Baseball Hit  & Baseball Base Hit - Double Plus & 87 \\
Baseball Idle  & Baseball Batter At Bat Idle & 45 \\
Baseball Pitching & Pitching A Baseball & 118 \\
Baseball Umpire & Umpire Calling A Strike & 79 \\
Capoeira & Capoeira Idle & 103 \\
Catwalk Sequence 04 & Female Sequence - Transition To Hands On Hips & 364 \\
Catwalk Walk & Model Walking Down The Catwalk & 454 \\
Crouched Walking & Crouched Walk With Rifle & 51 \\
Female Peek And Aim & Female Turnaround Gun Aim & 78 \\
Fireball & Street Fighter Hadouken & 101 \\
Focus & Shake Off Head Pain And Focus & 165 \\
Gangnam Style & The Popular K-Pop Dance & 371\\
Hip Hop Dancing & Female Hip Hop 'Rib Pops' Dancing & 101 \\
Hip Hop Dancing & Hip Hop Runningman Dance & 183 \\
Idle & Looking Over Both Shoulders & 120 \\
Kettlebell Swing & Russian Kettlebell Swing & 59 \\
Macarena Dance & Dancing The Macarena & 247 \\
Praying & Buckled Stand And Praying & 35 \\
Rumba Dancing & Female Rumba Dancing - Loop & 71 \\
Salsa Dancing & Female Salsa Dancing & 135 \\
Sit To Stand & Sitting To Standing & 68 \\
Sitting Clap & Sitting Unenthusiastic Clap & 195 \\
Sitting Disbelief & Sitting Disbelief With Hands On Head & 125 \\
Sitting Laughing & Sitting While Laughing & 250 \\
Start Walking & Standing To Start Walking With Rifle & 92 \\
Twist Dance & Doing The Twist Dance & 283 \\
Walking Backwards & One Foot At A Time In Combat Pose & 26 \\
Walking While Texting & Male Walking While Texting On A Smartphone & 120 \\
Yelling While Standing & Long Yell While Standing Leaning Back & 165 \\
\Xhline{4\arrayrulewidth}
\end{tabular}
% \vspace{-.1in}
% \caption{User study. We provide video comparisons of the two best performing methods to expert and non-expert animators. We ask them to choose the retargetted motion that is closer to the source motion and show the percentage of times each method is chosen.}
\caption{Test sequences used in this paper, search queries and number of frames per sequence. The sequences are downloaded for the characters Y bot, Kaya, Liam, Malcolm, Ortiz, Peasant Man and Warrok W Kurniawan.}
\vspace{-0.1in}
\label{table:test_data}
\end{table}

%% file: tables/skinny_comparison.tex
\begin{table}[h] 
% \small
\centering
\begin{tabular}{l||c|c||c|c}
\Xhline{4\arrayrulewidth}
\multicolumn{5}{c}{Skinny characters evaluation} \tabularnewline
\cline{1-5}
\multirow{3}{*}{ Method }  & \multicolumn{2}{c||}{Geometry evaluation} & \multicolumn{2}{c}{Motion evaluation} \tabularnewline
\cline{2-5} 
 & \makecell{ Inter-Penetrations} $\downarrow$ & \makecell{Vertex Contact \\ MSE}  $\downarrow$& \makecell{Foot Contact \\ Accuracy} $\uparrow$ & \makecell{Global Position \\ MSE} $\downarrow$\\
\Xhline{4\arrayrulewidth}
Ours & \textbf{0.49} & \textbf{2.24} & \textbf{0.98} & \textbf{0.31} \\
$E_{\mathit{full}} $ + IK & 0.82 & 2.81 & 0.81 & 1.97 \\
SAN \cite{aberman2020skeleton} & 1.21 & 4.50 & 0.61 & 1.23  \\
SAN + IK \cite{aberman2020skeleton} & 1.17 & 4.22 & 0.71 & 1.13  \\
PMnet \cite{Lim_pmnet_BMVC19} & 2.94 & 18.31 & 0.71 & 4.38  \\
NKN \cite{Villegas_2018_CVPR} & 2.11 & 13.00 & 0.72 & 13.95  \\
% HumanIK* & 6.59 & 5.11 & \textbf{0.98 / 0.97} & - \\
% HumanIK + Our Energy & - & - & - & - \\
\Xhline{4\arrayrulewidth}
\end{tabular}
% \vspace{-.1in}
\caption{Skinny characters evaluation. We evaluate the retargeted motion at the geometry and skeletal level. We evaluate the amount of self-penetrations, geometry contacts distance, foot contacts with the floor, and global joint positions in the retargeted motion.}
% \vspace{-0.1in}
\label{table:skinny}
\end{table}

%% file: tables/bulky_comparison.tex
\begin{table}[h] 
% \small
\centering
\begin{tabular}{l||c|c||c|c}
\Xhline{4\arrayrulewidth}
\multicolumn{5}{c}{Bulky characters evaluation} \tabularnewline
\cline{1-5}
\multirow{3}{*}{ Method }  & \multicolumn{2}{c||}{Geometry evaluation} & \multicolumn{2}{c}{Motion evaluation} \tabularnewline
\cline{2-5} 
 & \makecell{ Inter-Penetrations} $\downarrow$ & \makecell{Vertex Contact \\ MSE}  $\downarrow$& \makecell{Foot Contact \\ Accuracy} $\uparrow$ & \makecell{Global Position \\ MSE} $\downarrow$\\
\Xhline{4\arrayrulewidth}
Ours & \textbf{1.14} & \textbf{5.50} & \textbf{0.96} & 0.65 \\
$E_{\mathit{full}} $ + IK & 1.56 & 6.24 & 0.84 & 1.19 \\
SAN \cite{aberman2020skeleton} & 1.66 & 6.02 & 0.64 & 0.42  \\
SAN + IK \cite{aberman2020skeleton} & 1.48 & 7.34 & 0.74 & \textbf{0.35}  \\
PMnet \cite{Lim_pmnet_BMVC19} & 4.01 & 27.92 & 0.70 & 2.97  \\
NKN \cite{Villegas_2018_CVPR} & 4.29 & 16.72 & 0.70 & 2.34  \\
% HumanIK* & 6.59 & 5.11 & \textbf{0.98 / 0.97} & - \\
% HumanIK + Our Energy & - & - & - & - \\
\Xhline{4\arrayrulewidth}
\end{tabular}
% \vspace{-.1in}
\caption{Bulky characters evaluation. We evaluate the retargeted motion at the geometry and skeletal level. We evaluate the amount of self-penetrations, geometry contacts distance, foot contacts with the floor, and global joint positions in the retargeted motion.}
% \vspace{-0.1in}
\label{table:bulky}
\end{table}

%% file: figures/correspondences.tex
\begin{figure}[t!]
  \centering
  \includegraphics[width=1\linewidth]{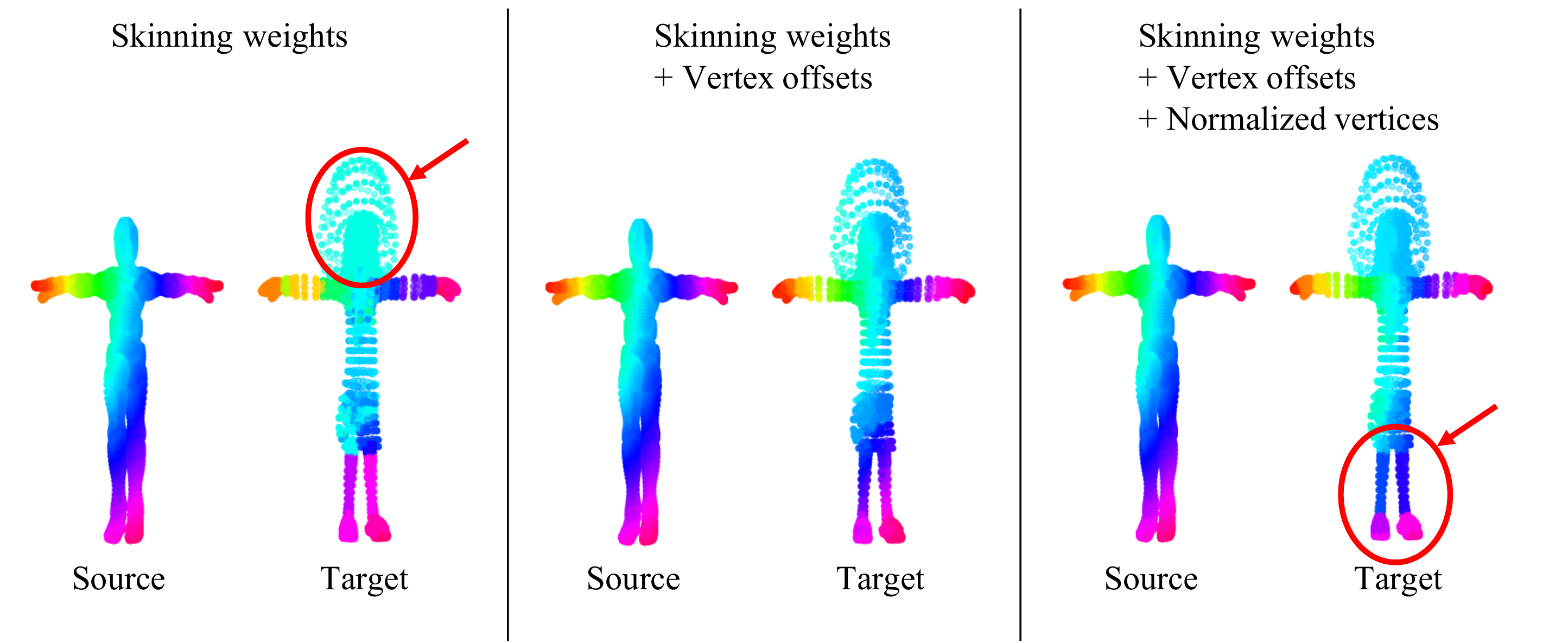}
  \caption{Correspondences experiments. We show color-coded results for using different features while computing vertex correspondences. We experimented with 1) only using skinning weights, 2) adding vertex offset from parent joint, and 3) adding normalized vertex coordinates. As pointed by the circles and arrows, feature combinations 1) and 3) result in inaccurate correspondence mapping due to ambiguity of skinning weights alone and noise added by the normalized vertices.}
  \label{fig:correspondence}
%   \vspace{-.1in}
\end{figure}